\documentclass[runningheads]{llncs}
\usepackage{graphicx}
\usepackage{amsmath,amssymb} 
\usepackage{color}
\usepackage{hyperref}

\usepackage{booktabs}
\usepackage{xcolor}
\usepackage[percent]{overpic}
\usepackage{wrapfig}

\def\showqualiresults{0} 

\newcommand{\refsec}[1]{Sec.~\ref{sec:#1}}

\newcommand{\reffig}[1]{Fig.~\ref{fig:#1}}
\newcommand{\reftab}[1]{Tab.~\ref{tab:#1}}

\newcommand{\knn}[0]{kNN}
\newcommand{\kmeans}[0]{K-means\;}

\definecolor{ceiling}{RGB}{0,255,0}
\definecolor{floor}{RGB}{0,0,255}
\definecolor{wall}{RGB}{0,255,255}
\definecolor{beam}{RGB}{255,255,0}
\definecolor{column}{RGB}{255,0,255}
\definecolor{window}{RGB}{100,100,255}
\definecolor{door}{RGB}{200,200,100}
\definecolor{table}{RGB}{170,120,200}
\definecolor{chair}{RGB}{255,0,0}
\definecolor{sofa}{RGB}{200,100,100}
\definecolor{bookcase}{RGB}{10,200,100}
\definecolor{board}{RGB}{200,200,200}
\definecolor{clutter}{RGB}{50,50,50}

\definecolor{Terrain}{RGB}{200, 90, 0}
\definecolor{Tree}{RGB}{0, 128, 50}
\definecolor{Vegetation}{RGB}{0, 220, 0}
\definecolor{Building}{RGB}{255, 0, 0}
\definecolor{Road}{RGB}{100, 100, 100}
\definecolor{GuardRail}{RGB}{200, 200, 200}
\definecolor{TrafficSign}{RGB}{255, 0, 255}
\definecolor{TrafficLight}{RGB}{255, 255, 0}
\definecolor{Pole}{RGB}{128, 0, 255}
\definecolor{Misc}{RGB}{255, 200, 150}
\definecolor{Truck}{RGB}{0, 128, 255}
\definecolor{Car}{RGB}{0, 200, 255}
\definecolor{Van}{RGB}{255, 128, 0}

\begin{document}
\title{Know What Your Neighbors Do: \\3D Semantic Segmentation of Point Clouds} 
\titlerunning{Know What Your Neighbors Do: 3D Semantic Segmentation of Point Clouds}
%
\author{
Francis Engelmann \and
Theodora Kontogianni \and
Jonas Schult \and
Bastian Leibe}
%
\authorrunning{F. Engelmann et al.}

\institute{
	RWTH Aachen University, Aachen, Germany.\\
	\email{\{engelmann, kontogianni, schult, leibe\}}@vision.rwth-aachen.de}
\maketitle              

\begin{figure}
\begin{center}
\includegraphics[width=0.83\linewidth]{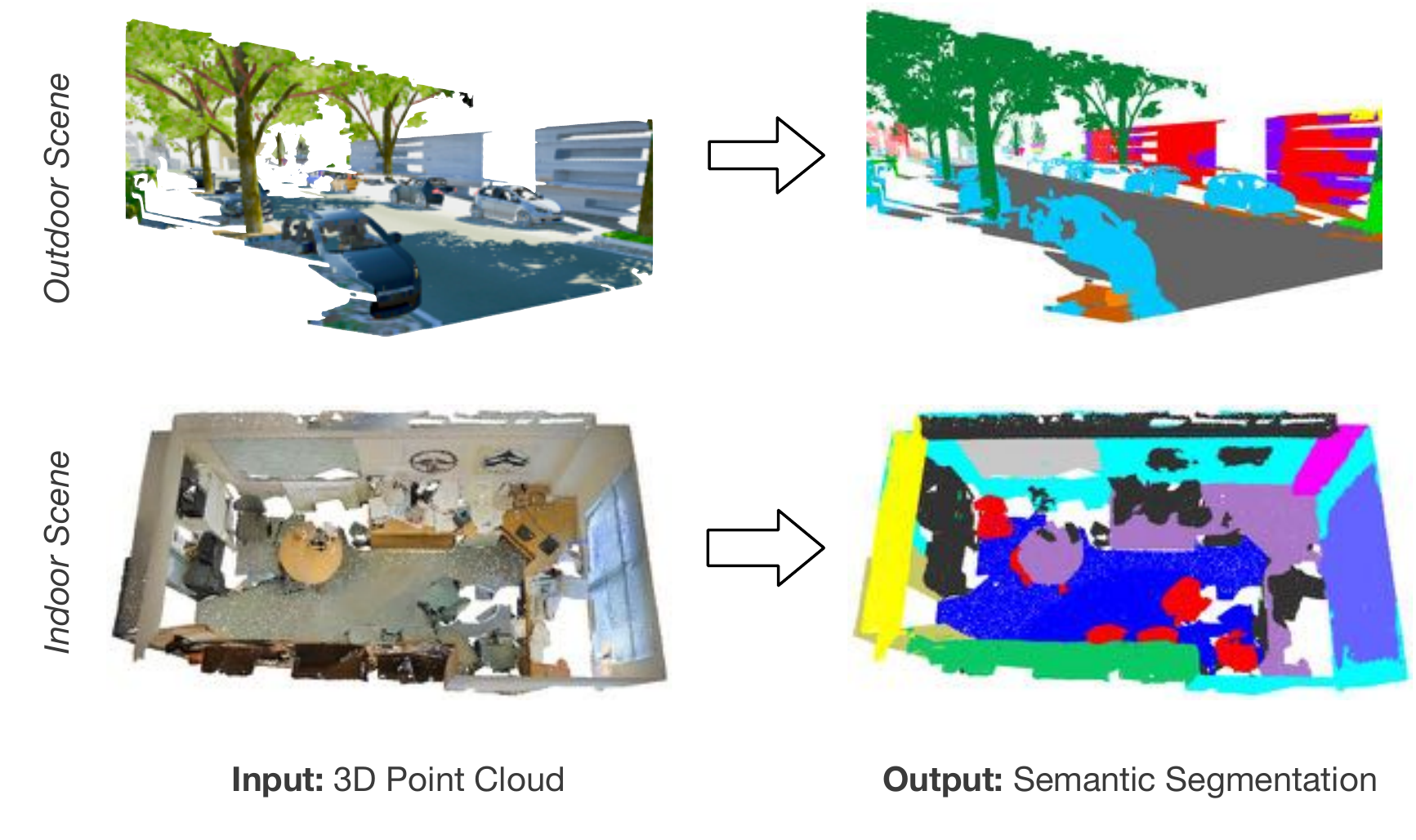} %
\end{center}
\caption{We present a deep learning framework that predicts a semantic label for each point in a given 3D point cloud. The main components of our approach are \emph{point neighborhoods} in different feature spaces and dedicated \emph{loss functions} which help to refine the learned feature spaces. 
\emph{Left}: point clouds from indoor and outdoor scenes. \emph{Right:} semantic segmentation results produced by the presented method.
}
\end{figure}

\begin{abstract}
In this paper, we present a deep learning architecture which addresses the problem of 3D semantic segmentation of unstructured point clouds.
Compared to previous work, we introduce grouping techniques which define \emph{point neighborhoods} in the initial world space and the learned feature space.
Neighborhoods are important as they allow to compute local or global point features depending on the spatial extend of the neighborhood.
Additionally, we incorporate dedicated loss functions to further structure the learned point feature space: the \emph{pairwise distance loss} and the \emph{centroid loss}.
We show how to apply these mechanisms to the task of 3D semantic segmentation of point clouds and report state-of-the-art performance on indoor and outdoor datasets.
\end{abstract}


\section{Introduction}
In the field of 3D scene understanding, semantic segmentation of 3D point clouds becomes increasingly relevant.
Point cloud analysis has found its application in indoor scene understanding and more recently has become an essential component of outdoor applications \cite{samp}.
This is due to the increasing availability and affordability of 3D sensors such as LiDAR or the Matterport scanner.

In the 2D image domain, for many tasks (including semantic segmentation) convolutional neural networks dominate the field.
2D convolutions allow processing large datasets with high resolution images by taking advantage of the locality of the convolutional operator.
They reduce the number of model parameters allowing for deeper and more complex models while being efficient
\cite{ChenPK0Y16,fullyConv,Noh_2015_ICCV,WuSH16a}.

However point clouds have no inherent order, such as pixel neighborhoods. 
They are generally sparse in 3D space and the density varies with the distance to the sensor.
Moreover, the number of points in a cloud can easily exceed the number of pixels in a high resolution image by multiple orders of magnitude.
All these properties make it difficult to process point clouds directly with traditional convolutional neural networks.

Recently, a lot of effort has been put into bridging the success from 2D scene understanding into the 3D world \cite{kd-networks,dec,vmv3d,segcloud,pointnet,pointnetpp,spg,urtasun,us}.
In this work, we aim to further narrow down the gap between 2D and 3D semantic scene understanding.
The straightforward approach of applying CNNs in the 3D space is implemented by preprocessing the point cloud into a voxel representation  first in order to apply 3D convolutions on that new representation \cite{vmv3d}. However 3D convolutions  have drawbacks. Memory and computational time grows cubicly on the number of voxels, restricting approaches to use coarse voxels grids. However, by doing so, one then introduces discretization artifacts (especially for thin structures) and loose geometric information such as point density.
Methods directly operating on the point cloud representation (e.g. \cite{pointnet,pointnetpp}) produce promising results.
However, in these methods, the point neighborhoods over which point features are aggregated are either global \cite{pointnet} or defined in a static coarse-to-fine approach \cite{pointnetpp}.  Either way, the inherent way of convolutions 
to capture the local structure has only been transferred in a limited fashion.

In this work, we propose to define neighborhoods in an adaptive manner that is primarily sensitive to the local geometry by using \kmeans clustering on the input point cloud features in the world space and secondly defining dynamic neighborhoods in the learned feature space using $k$ nearest neighbors (\knn).
Next comes the observation that a well structured feature space is essential for learning on point clouds.
Thus, we add dedicated loss functions which help shaping the feature space at multiple locations in the network.

We showcase the effectiveness of our approach on the task of semantic segmentation of 3D point clouds.
We present a comprehensive ablation study evaluating all introduced mechanisms.
We apply our method in different scenarios:
indoor data from the Stanford 3D Indoor Scene dataset \cite{s3dis} and ScanNet dataset \cite{scannet} as well as outdoor data from the Virtual KITTI 3D dataset \cite{us,vkitti}.


\section{Related Work}
\label{sec:related_work}
Before the introduction of deep learning methods, there have been numerous traditional approaches \cite{hackel2016fast,LaiICRA14,Munoz_3DPVT_2008,Xiong_ICRA_2011} applied to the task of semantically labelling 3D point clouds.
Since then, methods relying on deep learning can be roughly split into two groups: methods that impose structure on the unstructured 3D point cloud (by voxelization or projection) followed by standard convolutions, 
and methods that operate directly on the 3D point clouds:

\subsubsection{Voxelized Methods}
Up until recently, the standard method to perform semantic segmentation of 3D data involved \textit{voxelization}.
Voxelization approaches  transform the unstructured 3D point clouds into regular volumetric 3D grids (\textit{voxels}).
By doing so, 3D convolutions can  be directly applied to the voxels \cite{shapenet,voxnet}.
Alternatively, \emph{projection} approaches map the 3D points into 2D images as seen by virtual cameras.
Then, 2D convolutions are applied on the projections \cite{3dor,vmv3d}.
These methods suffer from major disadvantages: the mapping from a sparse representation to a dense one leads to an increased memory footprint.
Moreover, the fixed grid resolution results in discretization artifacts and loss of information. 
 
\subsubsection{Point Cloud Methods}
A new set of methods started with the work of PointNet\,\cite{pointnet}.
PointNet operates directly on 3D points.
The key idea is the extraction of point features through a sequence of MLPs processing the points individually (point features) followed by a max-pooling operation that describes the points globally (global features).
Point- and global-representations are fused (concatenation + MLP) before making the final label predictions.
Many methods followed the PointNet paradigm to operate directly on point clouds.
Where PointNet partitions the space into cuboidal blocks of fixed arbitrary size, others use octrees \cite{octree} or kd-trees \cite{kd-networks} to partition the space in a more meaningful way.
Furthermore, PointNet does not take into consideration the local geometry and surface information.
Clustering has been used in many classical approaches as a way of imposing structure, mostly as a prepossessing step \cite{Xiong_ICRA_2011,Xiong_3-dscene}.
So \cite{pointnetpp} and \cite{kd-networks} were introduced trying to apply hierarchical grouping of the points and incorporate local structure information.
The former used farthest point sampling and the latter kd-trees. The authors of \cite{dec} generalize the convolution operator on a spatial neighborhood.
Taking it further from local neighborhoods, \cite{spg} organizes the points into superpoints of homogeneous elements and defines relationships between them with the use of graph neural networks on their so-called superpoint graph.
In \cite{us} also cuboidal blocks are used, which act as superpoints and update their respective global features based on the surrounding blocks in space or scale using GRUs.

We now compare our method to the recent PointNet++ \cite{pointnetpp} which is an hierarchical extension of the original PointNet \cite{pointnetpp}.
PointNet (PN) globally aggregates point features using max-pooling.
PN++ forms local groups (or neighborhoods) based on the metric world space and collapses each group onto a single representative point.
This technique is repeated to increase the receptive field in each iteration.
In our work, we follow a similar approach by iteratively applying feature-space neighborhoods ($N_F$-modules, introduced later):
In every $\mathcal{N}_F$ iteration, each point is updated with the aggregated feature information from its \knn s.
Repeating this procedure allows the information to flow over many points, one hop per iteration.
Unlike \cite{pointnetpp}, we build the neighborhood based on the feature space, this allows the network to learn the grouping.
In PN++, neighborhoods are statically defined by a metric distance.

\subsubsection{Feature Networks} 
As a first step on our network, we learn strong features using a dedicated feature network.
The idea of extracting initial strong features is prominent in the field: In \cite{urtasun}, features are learned in 2D image space using CNN for the task of 2.5 semantic segmentation.
For the task of object detection, VoxelNet \cite{voxelnet} uses a cascade of PointNet like architectures named \emph{Voxel Feature Encoding} to obtain a more meaningful feature representation.


\section{Our Approach}
\label{sec:method}
In the following, we describe the main components of our network.
Starting from the initial point features (e.g. position and color), we learn more powerful feature representations using a new \emph{Feature Network} as described in \refsec{feature_network}.
Next, we define two kinds of neighborhoods (\refsec{feature_neighborhood}) within the point cloud, one defined on the learned feature space and one on the input world space.
Based on these groupings, we learn regional descriptors which we use to inform the feature points about their neighborhood. 
Finally, we further enforce structure on the learned feature space by defining two dedicated loss functions (\refsec{loss_pairwise}).


\subsection{Feature Network}
\label{sec:feature_network}

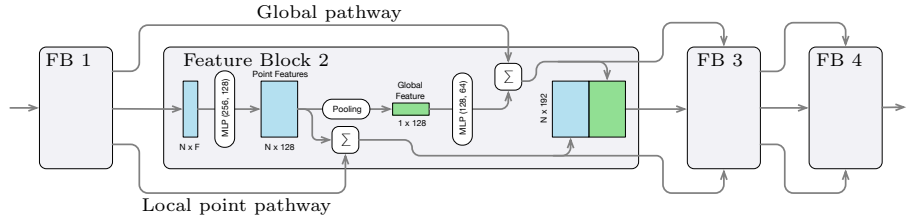
\begin{figure}[t!]
	\centering
	\begin{overpic}[width=1.0\textwidth, tics=10]{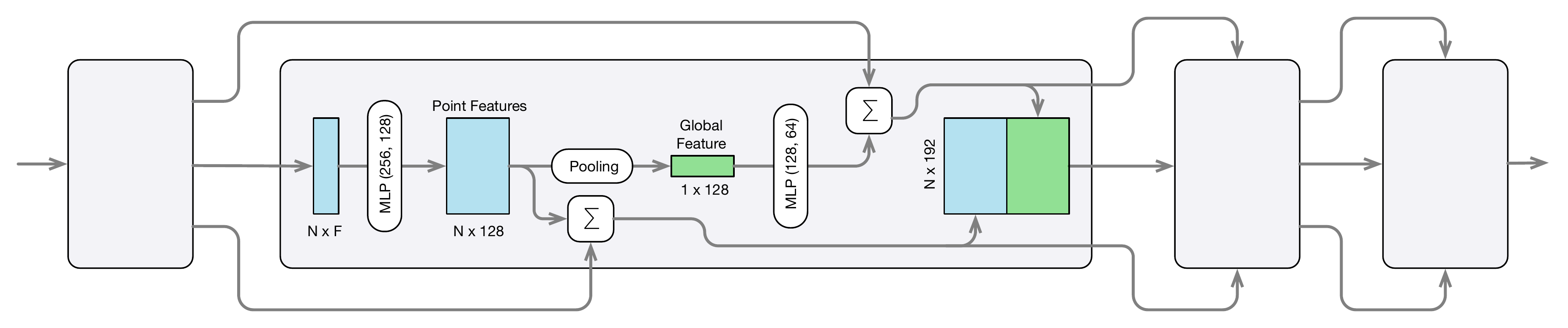}
	\put (5, 15) {\scriptsize{FB 1}}
	\put (20, 15) {\scriptsize{Feature Block 2}}
	\put (76, 15) {\scriptsize{FB 3}}
	\put (89, 15) {\scriptsize{FB 4}}
	\put (28, 20.2) {\scriptsize{Global pathway}}
	\put (15.5, -0.8) {\scriptsize{Local point pathway}}
	\end{overpic}
	\caption{
	The first component of our model is a \emph{feature network}, depicted above.
	It learns high-dimensional features using a series of \emph{feature blocks}.
	The details of a feature block are shown for feature block 2, the others are equal. 
	Further details and a motivation for the architecture are given in \refsec{feature_network}.
	Our complete model is shown in \reffig{architecture}.}
	\label{fig:feature_network}
\end{figure}

In this section, we describe our simple yet powerful architecture to learn point features. The goal of this component is to transform  input features - such as position and color - into stronger learned features.
It can be seen as the distilation of important elements from previous works, in particular \emph{PointNet}\,\cite{pointnet} and \emph{ConsolidationUnits}\,\cite{us}.
A schematic visualization is shown in \reffig{feature_network}.

The network is built from a sequence of \emph{feature blocks}.
Each feature block performs the following tasks:
Starting from a set of $N$ points, each  with feature dimension $F$, it produces refined \emph{point features} by passing the incoming features $F$ through a multi layer perceptron (MLP).
Then, a global representation is computed by aggregating all point features using max-pooling.
This \emph{global feature} is again passed through an MLP.
Finally, after vertically stacking the global feature $N$ times, it is concatenated with the point features.
Thus, a single feature block corresponds to a simplified PointNet. An important distinction is that feature blocks can be stacked to arbitrary depth.

In addition to the feature blocks, we introduce pathway connections which allow the individual feature blocks to consult features from previous layers.
We distinguish between the point features (local point pathway) and global features (global pathway).
Inspired by DenseNet\,\cite{densenet} and ResNet\,\cite{resnet}, these features can be combined either by concatenation or summation.
Our findings are that concatenation gives slightly inferior results over addition with the cost of a higher memory footprint.
At the same time, increasing the number of feature blocks in the network is even more important.
Thus, in the interest of scalability, in our final feature network we prefer addition over concatenation and use 17 \emph{feature blocks}.
Our experiments on different number of feature blocks and aggregation functions are summarized in \reftab{feature_network_eval}.
As a result, the feature network provides us with strong features required for the subsequent components.


\subsection{Neighborhoods}

We employ two different grouping mechanism to define neighborhoods over the point cloud:
The \emph{feature space neighborhood} $\mathcal{N}_F$ is obtained by computing the $k$ nearest neighbors (\knn) for each point in the learned feature space,
and the \emph{world space neighborhood}  $\mathcal{N}_W$ is obtained by clustering points using \kmeans in the world space. In this context, the world space corresponds to the features of the input point cloud, such as position and color.
In the following, we explain the two neighborhoods in more detail and show how they are used to update each point feature.

\begin{figure}[t!]
	\centering
	\begin{overpic}[width=1.0\textwidth, tics=10]{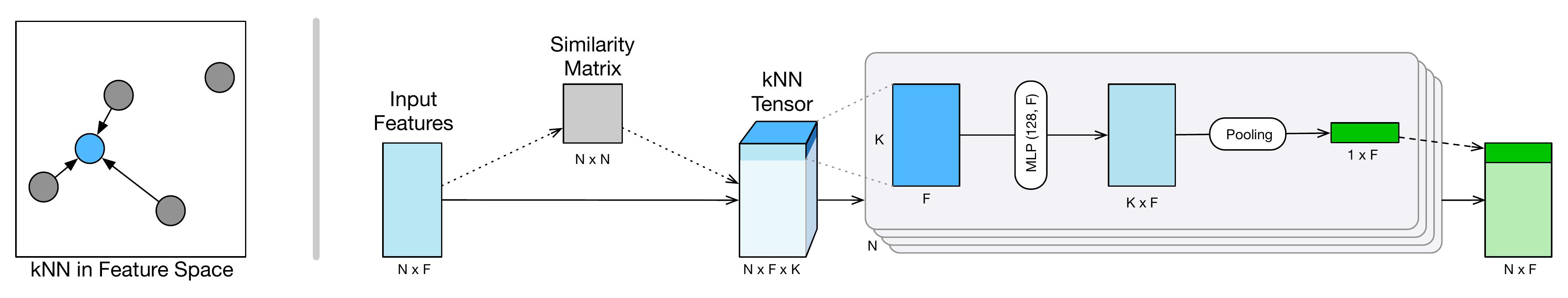}
	\end{overpic}
	\caption{
	The \emph{feature space neighborhood} $\mathcal{N}_F(\textbf{x})$ of a point $\textbf{x} \in \mathbb{R}^F$ in a $F$-dimensional feature space is defined as the \emph{k nearest neighbors} (\knn) in the feature space. \emph{Left:} Example for $k=3$. The point features \textbf{x} (blue) are updated based on the point features in the $\mathcal{N}_F$ neighborhood. \emph{Right:} Details of a $\mathcal{N}_F$-module for learning point features.}
	\label{fig:kmeans}
\end{figure}

\subsubsection{Feature space neighborhood $\mathcal{N}_F$} (See \reffig{kmeans})
\label{sec:feature_neighborhood}
From the set of $N$ input features of dimensionality $F$, we compute an $N \times N$ similarity matrix based on the pairwise $L_1$-distance between the feature points $\textbf{x}$. We concatenate the features of each point with the features of its $k$ nearest neighbors to construct a \knn\;tensor. Each slice in the tensor corresponds to an $\mathcal{N}_F$-neighborhood of a feature point $\textbf{x}_i$. Next, we learn a representation of this neighborhood using an MLP and we generate the updated feature point by applying max-pooling.
 This procedure is equivalent for each input feature and can be efficiently implemented using convolutions.
 We will refer to this architecture as an $\mathcal{N}_F$-module.

As such, an $\mathcal{N}_F$-module updates the local feature of a point based on its neighborhood in the feature space.
By concatenating multiple $\mathcal{N}_F$-modules, we can increase the receptive field of the operation, one hop at a time, which is comparable to applying multiple convolutions in the image space.

\subsubsection{World space neighborhood $\mathcal{N}_W$} 
\label{sec:world_neighborhood}

Unlike \knn, \kmeans assigns a variable number of points to a neighborhood.
\kmeans clustering is an iterative method, it alternatively assigns points to the nearest mean which represents the cluster center.
Then it recomputes the means based on the assigned points.
When applied to the world space, \kmeans can be seen as a pooling operation which reduces the input space and increases the receptive field by capturing long-range dependencies.
Additionally, we are offered a feature point representative per cluster by averaging over all cluster members in the feature space. 

We use this functionality in the $\mathcal{N}_W$-module:
we perform \kmeans clustering in the world space, and represent each cluster by the average over all feature points in the cluster.
Next, we concatenate this average to all the feature points within the same cluster.
We then again apply max-pooling which produces a regional descriptor for this cluster.
A visualization is shown in \reffig{architecture}.


\subsection{Loss Functions}
\label{sec:loss_functions}

In this section, we define the loss function $\mathcal{L}$ that is minimized during the training of our network. 
The classification loss $\mathcal{L}_{class}$ at the end of our network is realized as the cross entropy between predicted per-class probabilities and one-hot encoded ground truth semantic labels.
Beside the classification loss, we introduce two additional losses  $\mathcal{L}_{pair}$ and  $\mathcal{L}_{cent}$ which further help to shape the feature space.
The final loss is computed as the sum: $\mathcal{L} = \mathcal{L}_{class} + \mathcal{L}_{pair} + \mathcal{L}_{cent}$.

\begin{figure}[t!]
	\centering
	\begin{overpic}[width=1.0\textwidth, tics=10]{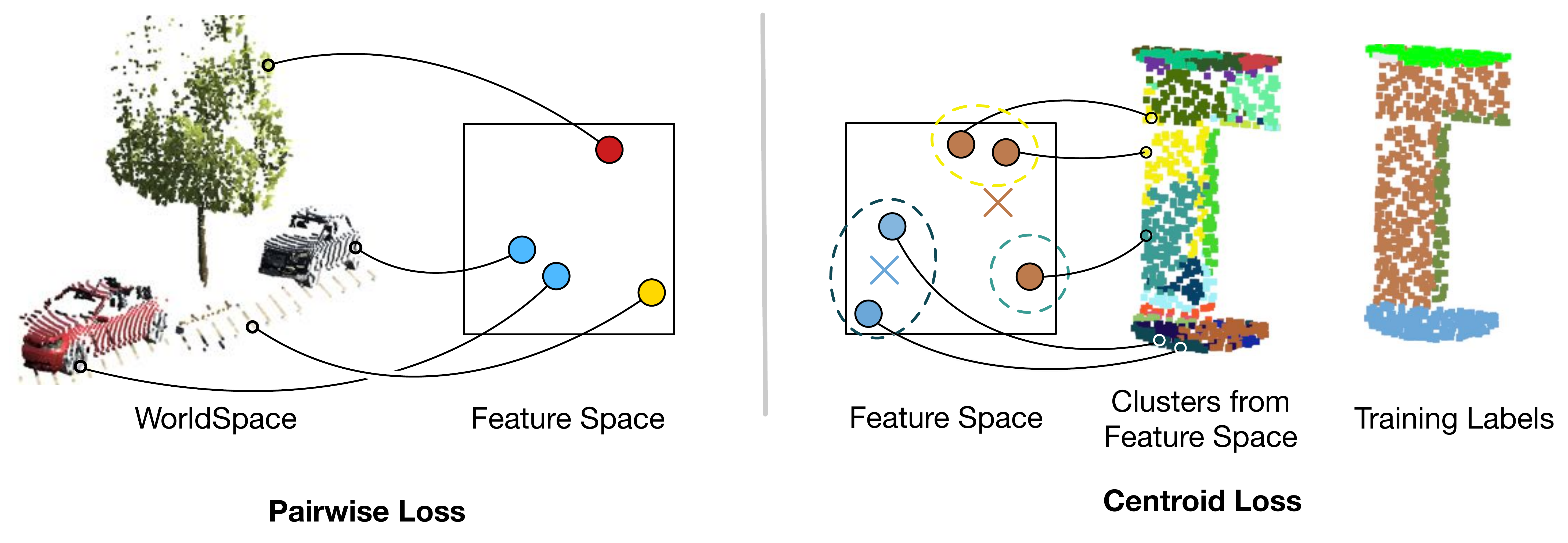}
	\end{overpic}
	\caption{\emph{Left:} The pairwise distance loss $\mathcal{L}_{pair}$ minimizes the distance in the feature space between points of the same semantic class while it increases the distance between points of different classes. \emph{Right:} The centroid loss $\mathcal{L}_{cent}$ minimizes the distance between features and their corresponding centroids, shown as crosses. The feature space is sketched as a 2D embedding. The point colors in the feature space represent training labels. To demonstrate the quality of our embedding, we further show clustering results (dashed lines) and their projection into world space (middle). See \refsec{loss_centroid} for details.}
	\label{fig:pairwise_loss}
\end{figure}

\subsubsection{Pairwise Similarity Loss $\mathcal{L}_{pair}$}
\label{sec:loss_pairwise}
So far, we assumed that points from the same semantic class are likely to be nearby in the feature space.
The \emph{pairwise similarity loss}, described in this section, explicitly enforces this assumption.
Similar to\,\cite{lecun_similarity}, we notice that semantic similarity can be measured directly as a distance in the feature space.
By minimizing pairwise distances, we can learn an embedding where two points sampled from the same object produce nearby points in the feature space.
Equivalently, two points originating from different objects have a large pairwise distance in the feature space.
This goal is illustrated with a 2D embedding in \reffig{pairwise_loss}.\\
\\
All we need is a pairwise distance matrix, which we already compute in the $\mathcal{N}_F$-module  (Sec. 3.2).
Hence, the distance loss is a natural extension of our network and comes at almost no additional memory cost.
For a pair of points $(i, j)$ with features $\textbf{x}_i$ and $\textbf{x}_j$, the loss is defined as follows:
\begin{equation}
     \ell_{i,j} =\left\{
                \begin{array}{ll}
                	\text{max}(|| \textbf{x}_i - \textbf{x}_j || - \tau_{\text{near}}, 0) & \text{if } \mathcal{C}_i = \mathcal{C}_j\\
                	\text{max}(\tau_{\text{far}} - || \textbf{x}_i - \textbf{x}_j ||, 0) & \text{if }  \mathcal{C}_i \not = \mathcal{C}_j
                \end{array}
              \right.
 \end{equation}
where $\tau_{\text{near}}$ and $\tau_{\text{far}}$ are threshold values and $\mathcal{C}_i$ is the semantic class of point $i$. 
Finally, the loss $\mathcal{L}_{pair}$ is computed as the sum over all pairwise losses $\ell_{i,j}$.

\subsubsection{Centroid Loss $\mathcal{L}_{cent}$}
\label{sec:loss_centroid}
This loss reduces the within-class distance by minimizing the distance between point features $\textbf{x}_i$ and a corresponding representative feature $\overline{ \textbf{x}}_i$ (centroid). It makes the features in the feature space more compact.
During training, the representative feature can be computed as the mean feature over all points from the same semantic class.
An illustration is shown in \reffig{pairwise_loss}.
We define the centroid loss as the sum over all $(\textbf{x}_i$, $\overline{ \textbf{x}}_i)$ pairs:

\begin{equation}
     \mathcal{L}_{cent} = \underset{i \in {[1..N]}}{\sum} || \textbf{x}_i -  \overline{ \textbf{x}}_i ||
 \end{equation}
 where ${N}$ is the total number of points.
As distance measure $|| \cdot ||$, we found the cosine distance to be more effective than the $L_1$ or $L_2$ distance measures.


\begin{figure}[t!]
	\centering
	\begin{overpic}[width=1\textwidth, tics=10]{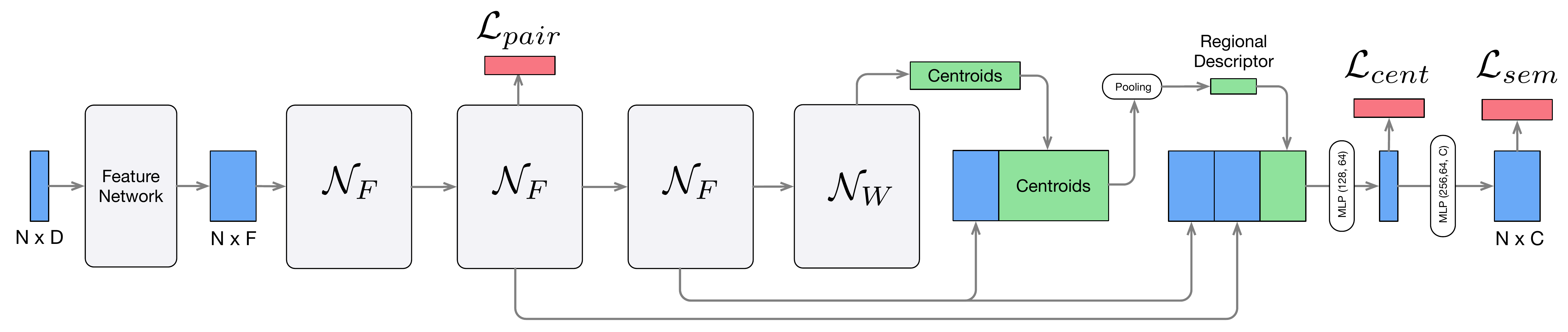}
	\end{overpic}
	\caption{Our complete network architecture. It consists of a \emph{Feature Network} (\refsec{feature_network}), followed by three $\mathcal{N}_F$-modules and one $\mathcal{F}_W$-module (\refsec{world_neighborhood}). Point features are represented by blue rectangles, losses are shown in red (\refsec{loss_functions}). Green blocks represent features computed over clusters in the world space.}
	\label{fig:architecture}
\end{figure}

\section{Implementation Details}
\label{sec:implementation}
In this section, we describe the integration of the aforementioned components into a deep learning architecture.
The complete model is depicted in \reffig{architecture}. We start by learning strong features using our \emph{Feature Network}\,(\refsec{feature_network}) producing $F$-dimensional features. See \reftab{feature_network_eval} for an evaluation and discussion on the architecture.
We then feed these features into three stacked $\mathcal{N}_F$-modules. The subsequent $\mathcal{N}_W$-module computes a regional descriptors for each cluster (based on world space with descriptors form the feature space). We concatenate the regional descriptors to its corresponding feature points of the second and third $\mathcal{N}_F$-module. The concatenated features are passed through another MLP after which we compute the centroid loss. Finally, we reduce the feature points to 13 dimensions corresponding to the number of semantic classes in our datasets.
The pairwise distance loss is computed in the beginning in the network.
This informs the networks as early as possible which points should have similar features.
This provides early layers a stronger signal of what should be learned and simplifies gradient propagation as the gradient is passed through fewer layers \cite{segaware}.
Although the distance loss could be appended at each point where a similarity matrix is computed, we found it most effective to add it to the second $\mathcal{N}_F$-module. An ablation study is provided in \reftab{ablation_comp} and shows the contribution of each component in the performance.

\begin{table}[t]
\caption{
Ablation study highlighting the contribution of the individual components of our pipeline.
The reported numbers refer to our validation set}
\center
\begin{tabular}[b]{r ccc}
\toprule
 Components & \, oAcc \, & \, mAcc \, & \, mIoU \, \\
\midrule
Feature Network (FN) (Sec. \ref{sec:feature_network})		& 82.43 & 56.96 & 47.25 \\ 
$\mathcal{N}_F$*3 (\refsec{feature_neighborhood})	& 81.70 & 55.14 & 47.37 \\
$\mathcal{N}_F$*3 + $\mathcal{L}_{pair}$ (\refsec{loss_pairwise}) & 82.51 & 58.20 & 49.41 \\ 
FN + $\mathcal{N}_F$*3 +  $\mathcal{L}_{pair}$ & 83.84 & 58.29 & 50.56 \\
FN + $\mathcal{N}_F$*3 + $\mathcal{N}_W$ + $\mathcal{L}_{pair}$ (\refsec{world_neighborhood}) & 84.31 & 59.18 & 50.95 \\
FN + $\mathcal{N}_F$*3 + $\mathcal{N}_W$ + $\mathcal{L}_{pair}$ + $\mathcal{L}_{cent}$ (\refsec{loss_centroid})	& 84.19 & 60.59 & 51.56 \\ 
\bottomrule
\end{tabular}
\label{tab:ablation_comp}
\end{table}


\section{Evaluation}
\label{sec:evaluation}
In this section, we evaluate the performance of our approach on the task of 3D semantic segmentation. 
We show qualitative and quantitative results and compare them to previous methods.
We evaluate our method on multiple datasets: two indoor and one outdoor dataset showing the versatility of our approach.
For each dataset, we report the \emph{overall accuracy} (oAcc), the \emph{mean class accuracy} (mAcc) and the \emph{mean class intersection-over-union} (mIoU).

\subsection{Indoor Datasets}
\begin{figure}
\centering
\colorbox{ceiling}{\scriptsize\strut Ceiling}%
\colorbox{floor}{\scriptsize\strut \textcolor{white}{Floor}}%
\colorbox{wall}{\scriptsize\strut Wall}%
\colorbox{beam}{\scriptsize\strut Beam}%
\colorbox{column}{\scriptsize\strut Column}%
\colorbox{window}{\scriptsize\strut \textcolor{white}{Window}}%
\colorbox{door}{\scriptsize\strut Door}%
\colorbox{table}{\scriptsize\strut Table}%
\colorbox{chair}{\scriptsize\strut Chair}%
\colorbox{sofa}{\scriptsize\strut Sofa}%
\colorbox{bookcase}{\scriptsize\strut Bookcase}%
\colorbox{board}{\scriptsize\strut Board}%
\colorbox{clutter}{\scriptsize\strut \textcolor{white}{Clutter}}%

\ifx\showqualiresults\undefined
\else
\includegraphics[width=0.24\linewidth]{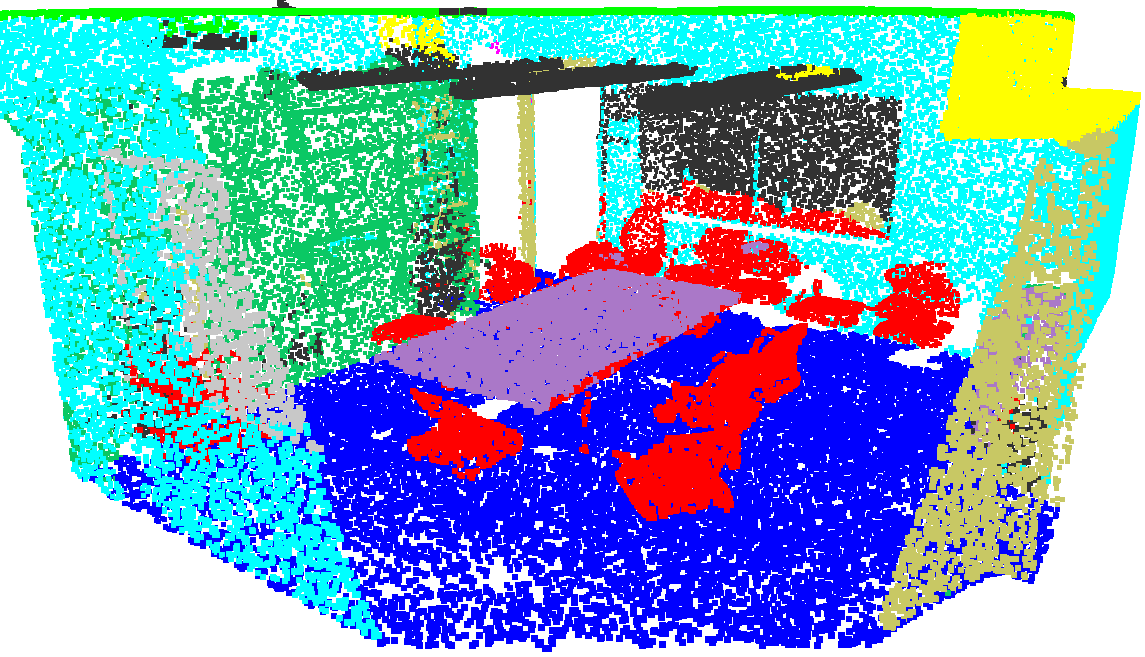} %
\includegraphics[width=0.24\linewidth]{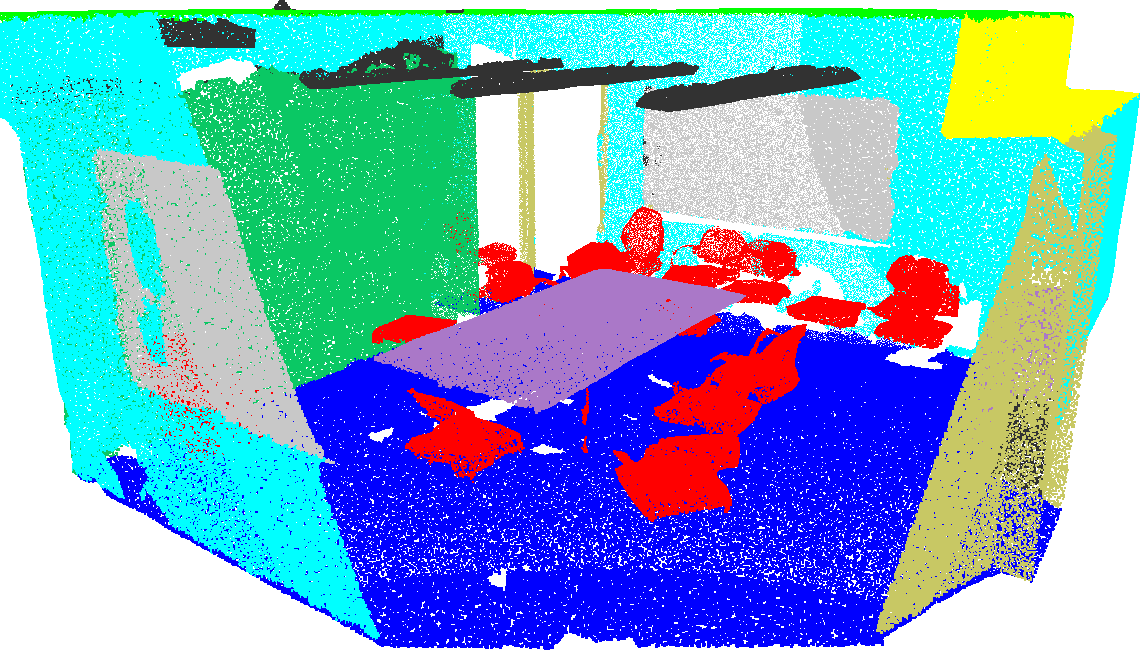} %
\includegraphics[width=0.24\linewidth]{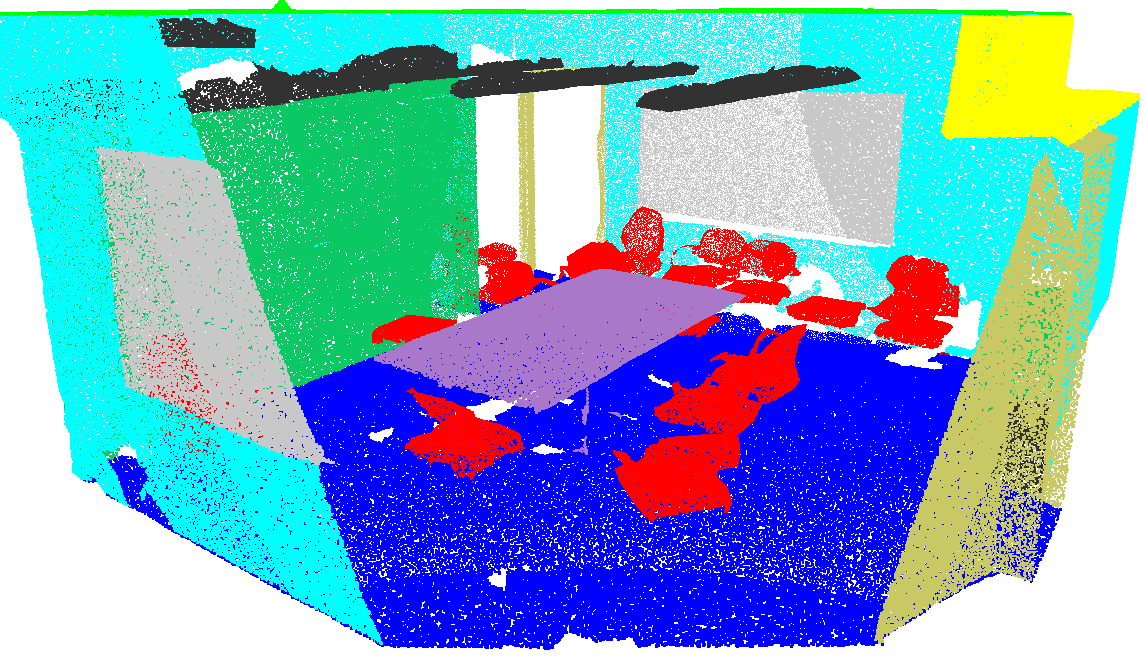} %
\includegraphics[width=0.24\linewidth]{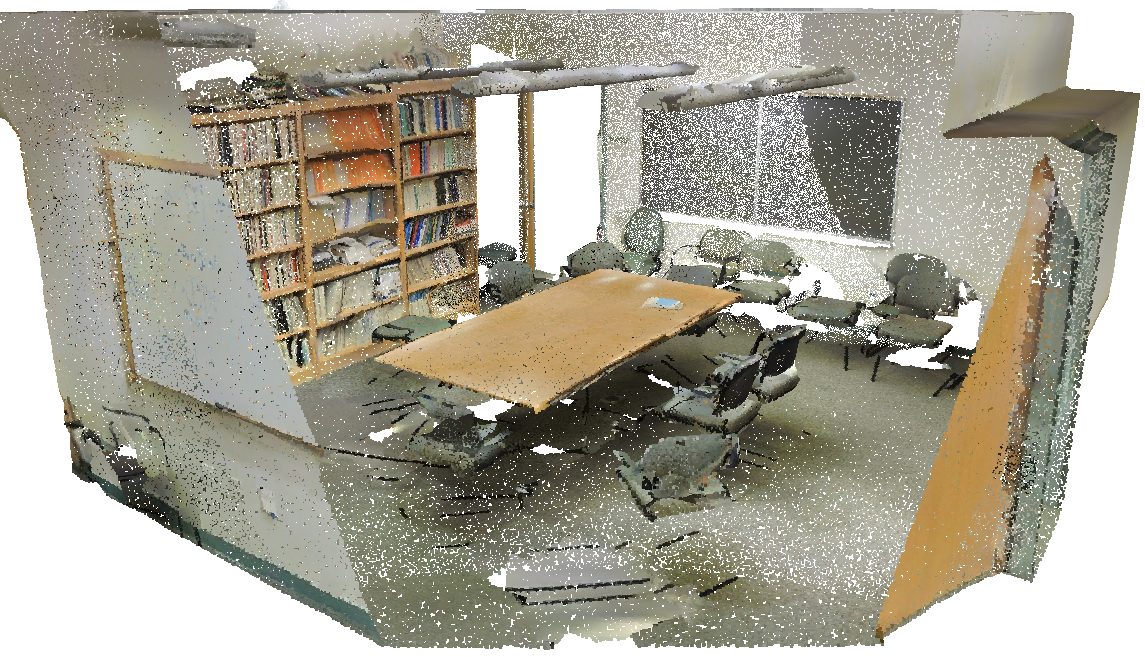} %

\includegraphics[width=0.24\linewidth]{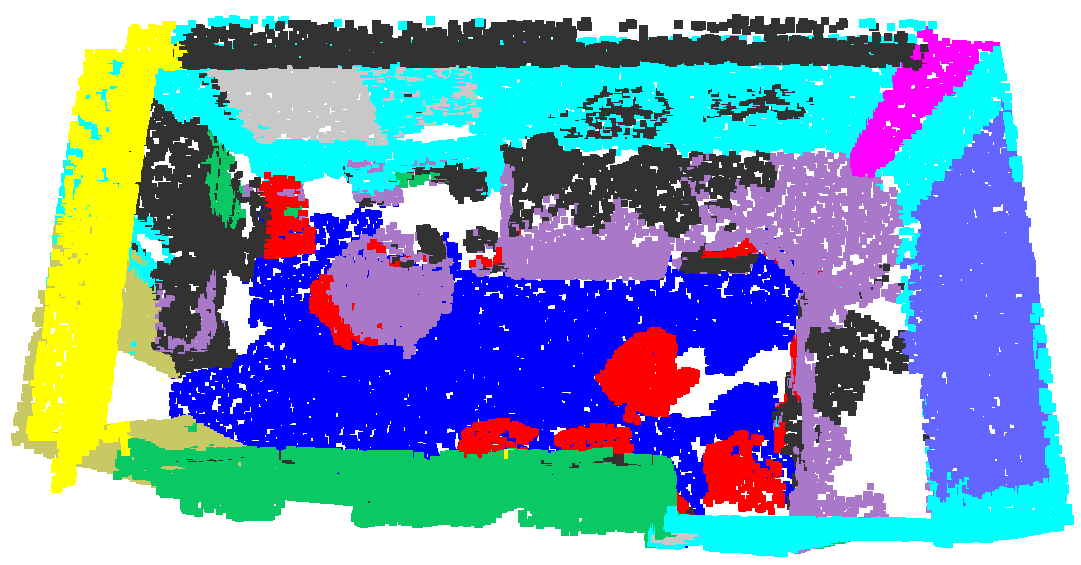} %
\includegraphics[width=0.24\linewidth]{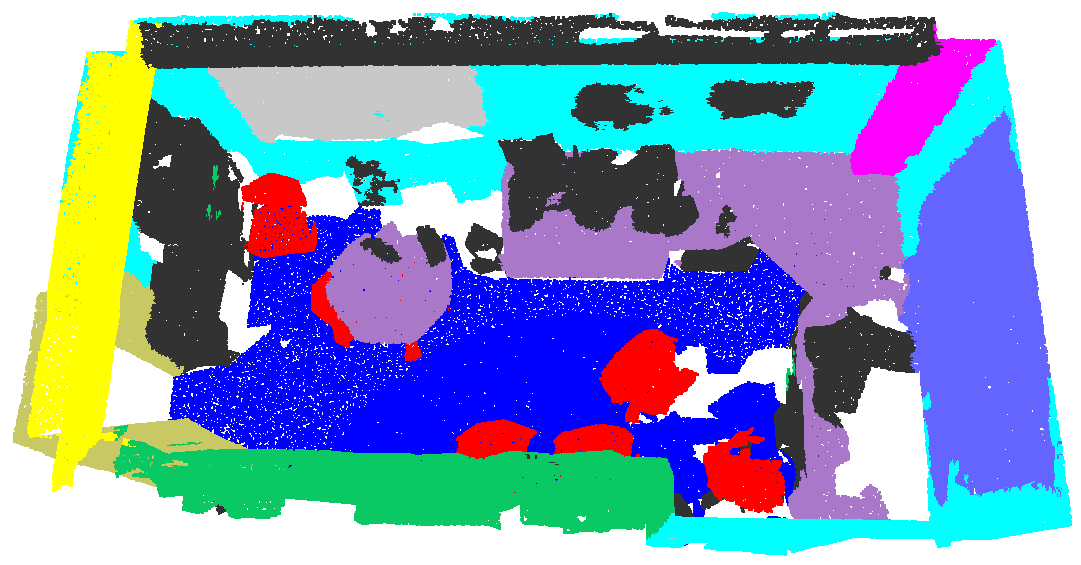} %
\includegraphics[width=0.24\linewidth]{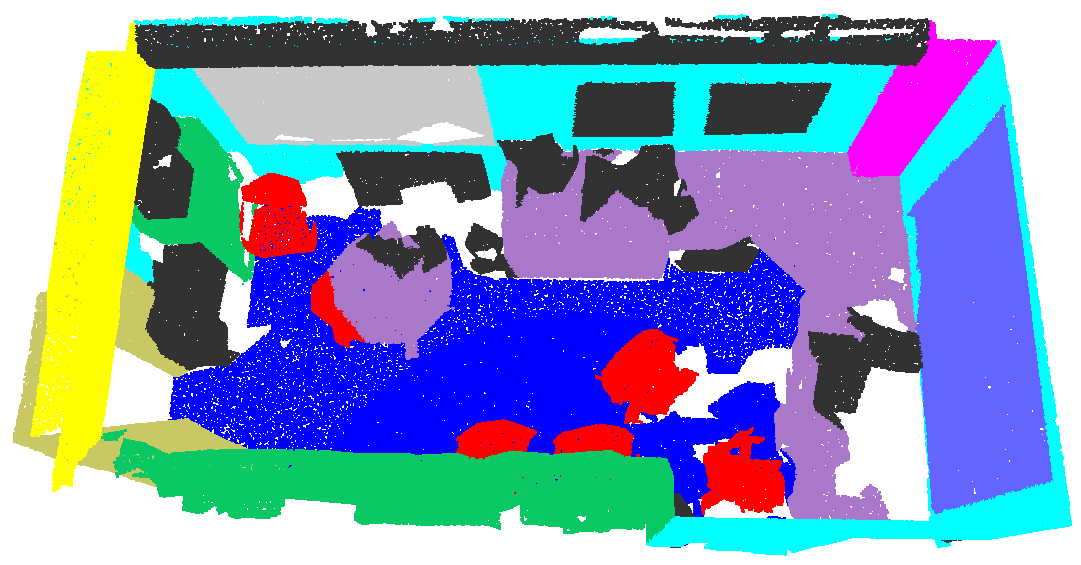} %
\includegraphics[width=0.24\linewidth]{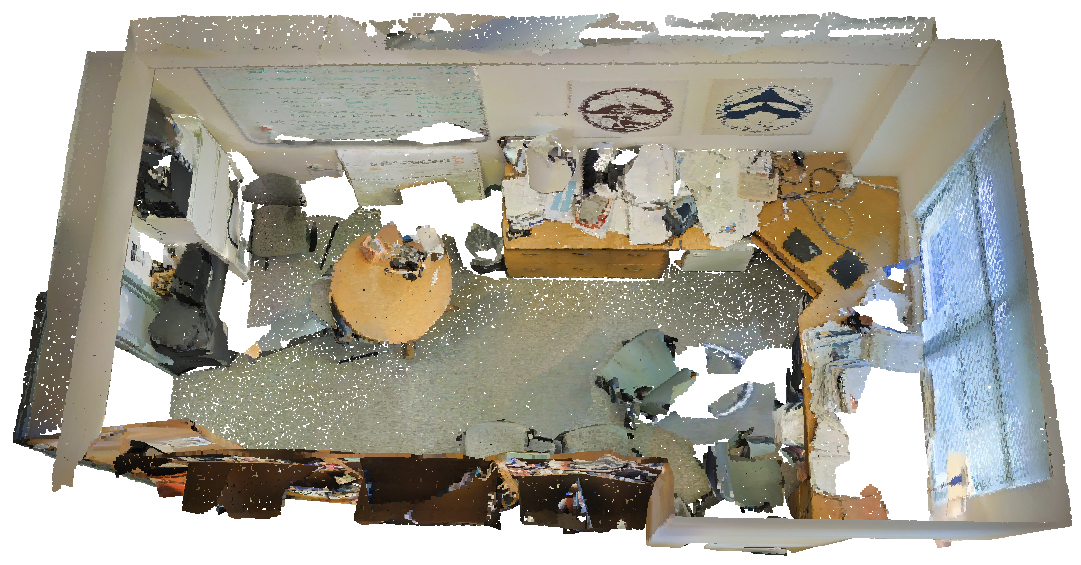} %

\includegraphics[width=0.24\linewidth]{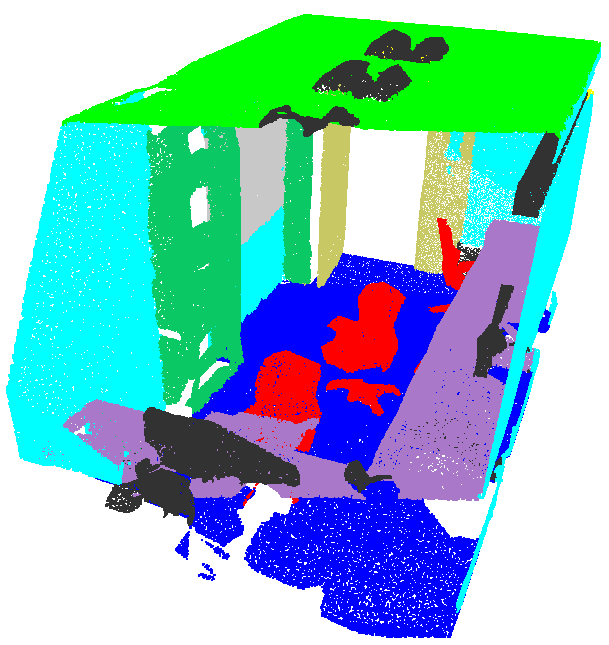} %
\includegraphics[width=0.24\linewidth]{images/s3dis/area_6/area_6_office_23_pred.png} %
\includegraphics[width=0.24\linewidth]{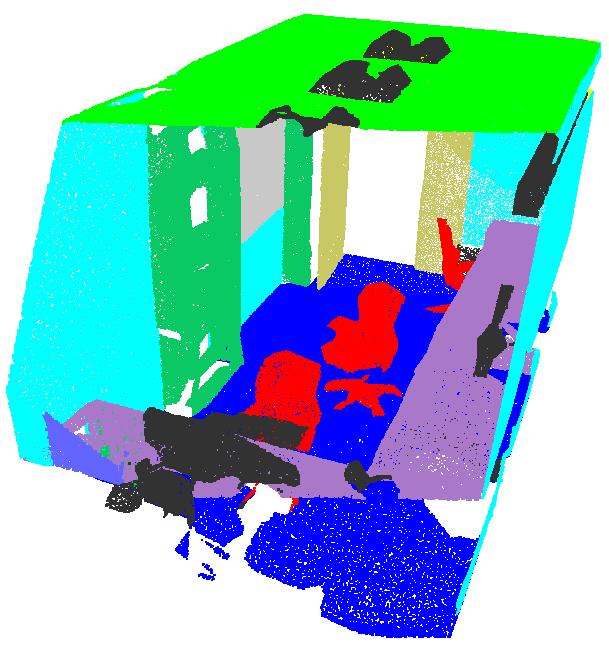} %
\includegraphics[width=0.24\linewidth]{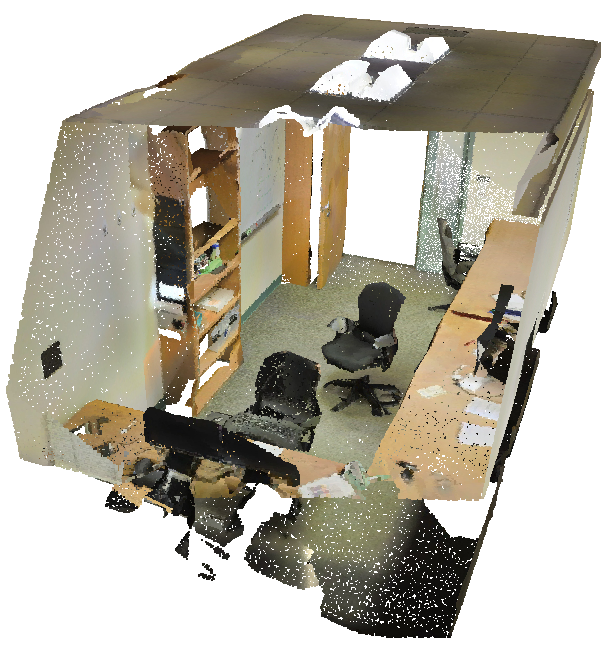} %

\includegraphics[width=0.24\linewidth]{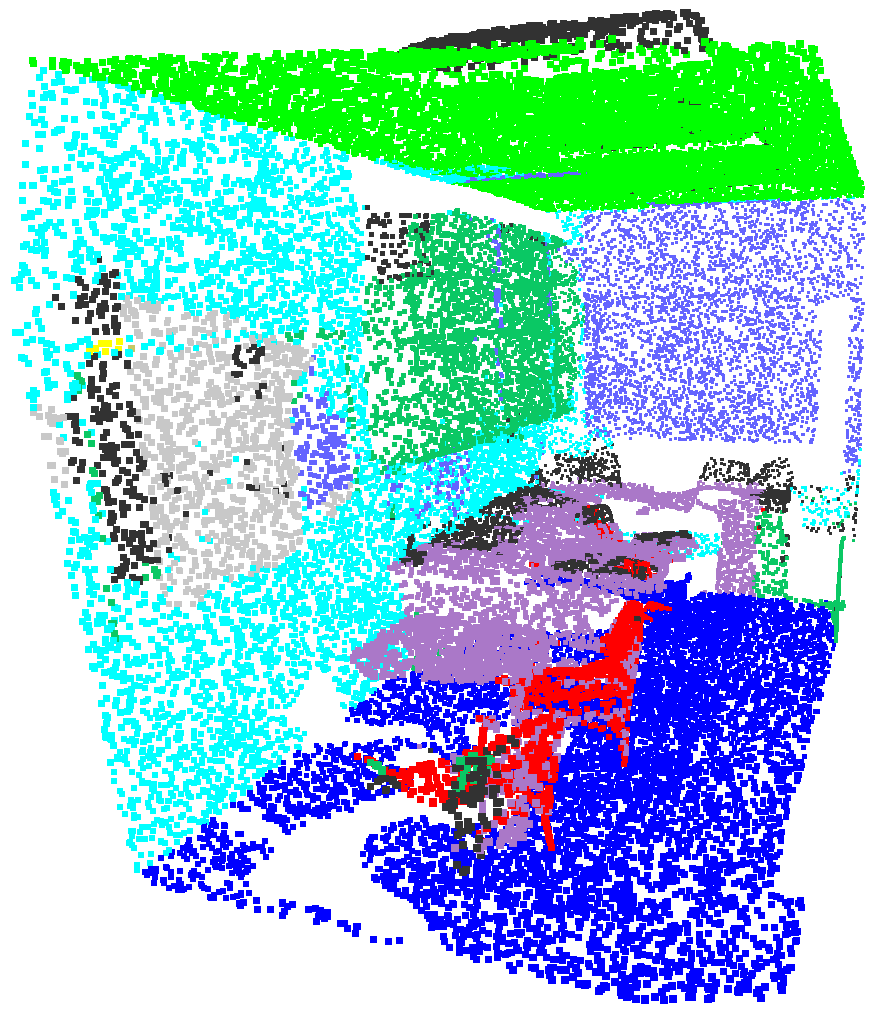} %
\includegraphics[width=0.24\linewidth]{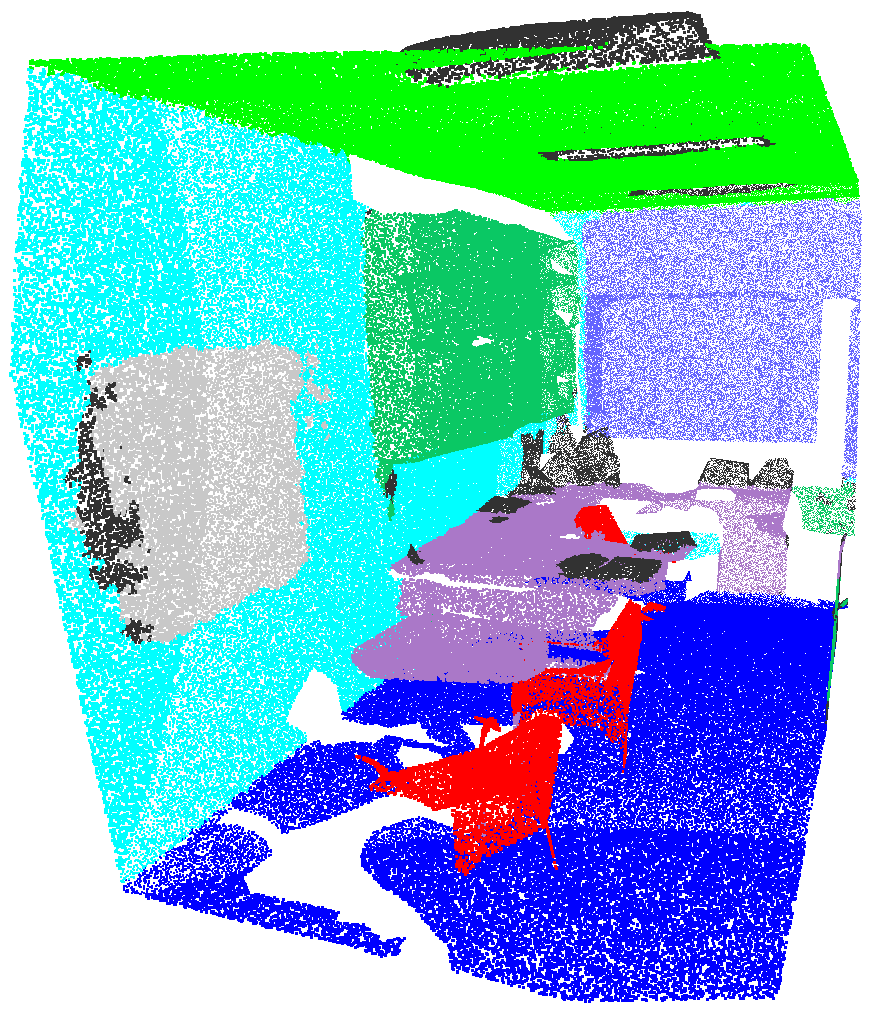} %
\includegraphics[width=0.24\linewidth]{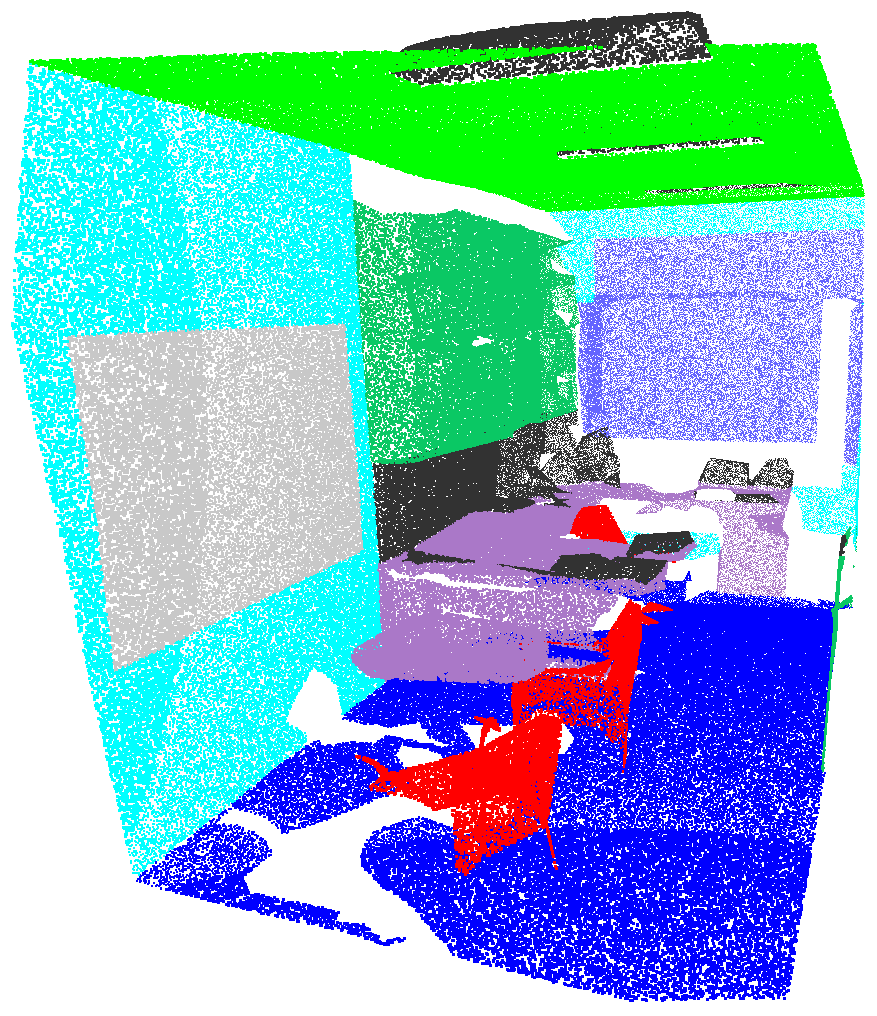} %
\includegraphics[width=0.24\linewidth]{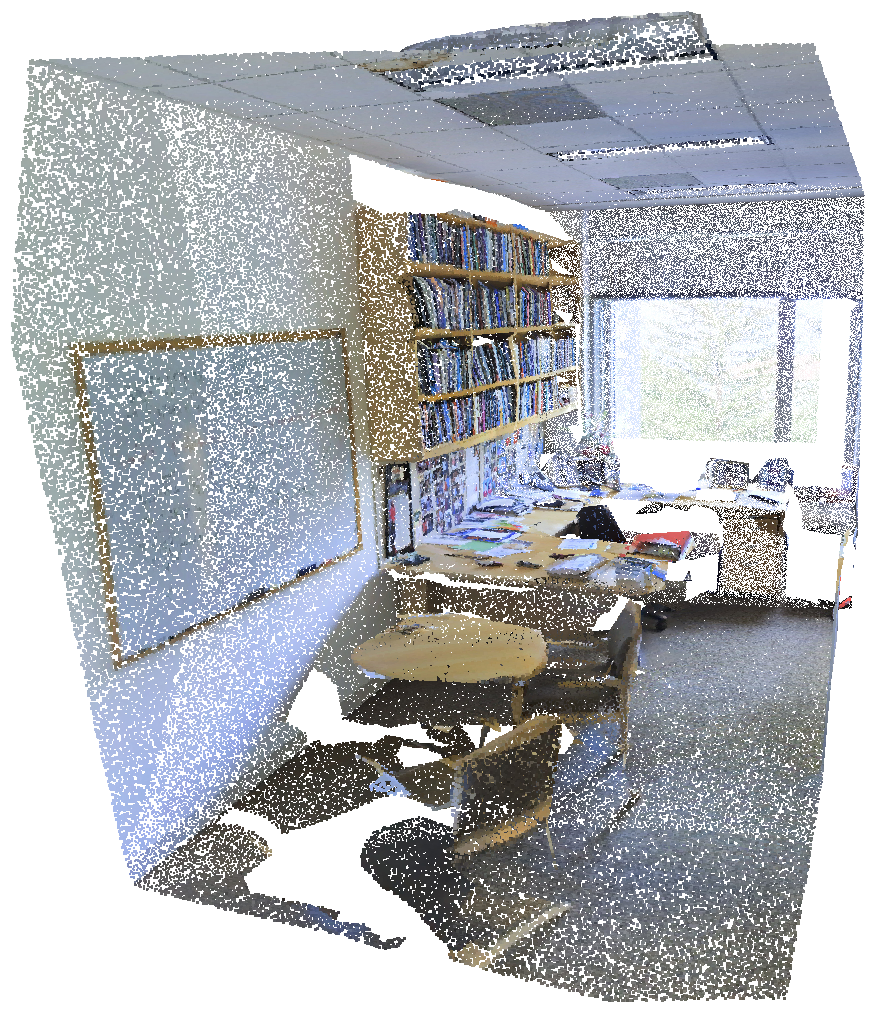} %
\fi

\begin{minipage}{0.24\textwidth}
\center
\textbf{\small Pointnet}
\label{fig:no_color}
\end{minipage}
\begin{minipage}{0.24\textwidth}
\center
\textbf{\small Ours}
\label{fig:color}
\end{minipage}
\begin{minipage}{0.24\textwidth}
\center
\textbf{\small Ground Truth}
\label{fig:gt}
\end{minipage}
\begin{minipage}{0.24\textwidth}
\center
\textbf{\small RGB}\label{fig:rgb}
\end{minipage}
\caption{
	\label{fig:quali_results_s3dis}
	Qualitative results on S3DIS dataset. We show three exemplary rooms. Our method provides segmentations of objects with minimal noise and clear boundaries. As pointed out in the qualitative results, our method performs quite well in challenging objects like 'board' and 'bookcase'.}
\end{figure}

We evaluate our model on the \textit{Stanford Large-Scale 3D Indoor Spaces} (S3DIS) dataset\,\cite{s3dis} and the \emph{ScanNet} dataset\,\cite{scannet}. Both datasets have recently become popular to evaluate 3D semantic segmentation methods. The S3DIS dataset consists of 6 different indoor areas, totaling to 272 rooms. Each point is labeled as one of 13 semantic classes as shown in \reffig{quali_results_s3dis}.
The ScanNet dataset contains 1523 RGB-D scans labeled with 20 different semantic classes.

\subsection{Outdoor Dataset}
We apply our approach to the vKITTI3D dataset, a large-scale outdoor data set in an autonomous driving setting. Introduced in\,\cite{us}, this dataset is an adaptation of the synthetic \emph{Virtual KITTI} dataset\,\cite{vkitti} for the task of semantic segmentation of 3D point clouds. It is split in 6 different sequences containing 13 semantic classes listed in \reffig{quali_results_vkitti3d}. 

\begin{table}[t]
\caption{
\textbf{Stanford Large-Scale 3D Indoor Spaces.}
6-fold cross validation results.
We can present state-of-the-art results in the more difficult CV and slightly inferior results on Area 5.
}
\center
\begin{tabular}{rccc}
\toprule
\textbf{S3DIS\,\cite{s3dis}: 6-fold CV} & \, oAcc \, & \, mAcc \, & \, mIoU \, \\
\midrule
PointNet\,\cite{pointnet} & 78.62 & - & 47.71\\
G+RCU\,\cite{us}           & 81.1 &  66.4 & 49.7 \\
DGCNN \,\cite{dgcnn}	    & 84.1 & - & 56.1\\
PointNet++\,\cite{pointnetpp}  &  81.03  & 67.05 & 54.49\\ 
RSN\,\cite{rsn}					& - & 66.45 & 56.47\\ 
SPG\,\cite{spg}              & \textbf{85.5} & \textbf{73.0} & \textbf{62.1}\\
Ours & 83.95 & 67.77 & 58.27 \\
\bottomrule
\end{tabular}
\label{tab:quanti_s3dis_cv}
\end{table}
\begin{table}[th]
\caption{
\textbf{Stanford Large-Scale 3D Indoor Spaces.} Results on Area 5}
\center
\begin{tabular}{r ccc}
\toprule
\textbf{S3DIS\,\cite{s3dis}: Area 5} & \, oAcc \, & \, mAcc \, & \, mIoU \, \\
\midrule
PointNet\,\cite{pointnet}      & -        & 48.98 & 41.09\\
MS + CU(2)\,\cite{us}  & - & 52.11 & 43.02 \\
G + RCU\,\cite{us} & - & 54.06 & 45.14 \\
SEGCloud\,\cite{segcloud} & -        & 57.35 & 48.92\\
SPG\,\cite{spg}                   & \textbf{86.38} & \textbf{66.50} & \textbf{58.04}\\
\midrule
Ours & 84.15 & 59.10 & 52.17 \\
\bottomrule
\end{tabular}
\label{tab:quanti_s3dis_a5}
\end{table}

\subsection{Training Details}
For the experiments on the S3DIS dataset, we follow a similar training procedure as \cite{pointnet} i.e. we split the rooms in blocks  of 1 m$^2 $ on the ground plane.
From each block we randomly sample 4096 points.
During evaluation, we predict class labels for all points. 
Additionally, we add translation augmentation to the block positions.
Each point is represented by a 9D feature vector $[x, y, z, r, g, b, x', y', z']$ consisting of the position $[x, y, z]$, color $[r, g, b]$ and normalized coordinates $[x', y', z']$ as  in  \cite{pointnet}.
The hyperparameters of our method are set as follows: 
for the \knn-clustering, we set $k=30$ and use the $L_1$-distance measure.
For \kmeans we dynamically set $\text{K} = \lfloor \text{N} / 52\rfloor$ where N is the number of points per block. 
We report scores on  a 6-fold cross validation across all areas in \reftab{quanti_s3dis_cv} along with the detailed scores of per class IoU  in \reftab{s3dis_color_iou}. Additionally, we provide scores for Area 5 in  \reftab{quanti_s3dis_a5} to compare ourself to \cite{segcloud} and \cite{spg}.

On the ScanNet dataset\,\cite{scannet}, we use the reference implementation of PointNet++ to train and evaluate our model.
This approach allows us to focus on the comparison of the models while abstracting from the training procedures. All hyperparameters remain the same. The results are shown in \reftab{quanti_scannet}

\begin{table*}[t]
\caption{IoU per semantic class on the S3DIS dataset.
We compare our model against the original PointNet and other recent methods.
On average our method outperforms the current state-of-the-art by a large margin, specifically on 'bookcase' and 'board' while being slightly worse on 'beam' and 'sofa'
}
\begin{center}
\begin{scriptsize}
\begin{tabular}[t]{r|c|ccccccccccccc}
\textbf{Method} & mIoU &
\rotatebox{90}{Ceiling}&%
\rotatebox{90}{Floor}&%
\rotatebox{90}{Wall}&%
\rotatebox{90}{Beam}&%
\rotatebox{90}{Column}&%
\rotatebox{90}{Window}&%
\rotatebox{90}{Door}&%
\rotatebox{90}{Table}&%
\rotatebox{90}{Chair}&%
\rotatebox{90}{Sofa}&%
\rotatebox{90}{Bookcase}&%
\rotatebox{90}{Board}&%
\rotatebox{90}{Clutter}\\
\midrule
PointNet \cite{pointnet} & 47.6 &
88.0 &
88.7 &
69.3 &
42.4 &
23.1 &
47.5 &
51.6 &
54.1 &
42.0 &
9.6 &
38.2 &
29.4 &
35.2 \\
MS+CU(2) \cite{us} & 47.8 &
88.6 &
95.8 &
67.3 &
36.9 &
24.9 &
48.6 &
52.3 &
51.9 &
45.1 &
10.6 &
36.8 &
24.7 &
37.5 \\
SegCloud\cite{segcloud} & 48.9  &
90.1 &
\textbf{96.1} &
69.9 &
0.0 &
18.4 &
38.4 &
23.1 &
\textbf{75.9} &
\textbf{70.4} &
58.4 &
40.9 &
13.0 &
42.0 \\
G+RCU \cite{us} & 49.7 &
90.3 &
92.1 &
67.9 &
44.7 &
24.2 &
52.3 &
51.2 &
58.1 &
47.4 &
6.9 &
39.0 &
30.0 &
41.9 \\
SPG \cite{spg} & \textbf{62.1} &
89.9  &
95.1  &
76.4  &
\textbf{62.8}  &
\textbf{47.1}  &
\textbf{55.3}  &
\textbf{68.4}  &
73.5  &
69.2  &
\textbf{63.2}  &
45.9  &
8.7  &
52.9 \\
\midrule
Ours & 58.3 &
\textbf{92.1}  &
90.4  &
78.5  &
37.8  &
35.7  &
51.2  &
65.4  &
64.0  &
61.6  &
25.6  &
\textbf{51.6}  &
\textbf{49.9}  &
\textbf{53.7} \\
\bottomrule
\end{tabular}
\end{scriptsize}
\end{center}
\label{tab:s3dis_color_iou}
\end{table*}

\begin{figure*}
\centering
\colorbox{Terrain}{\tiny \strut Terrain}%
\colorbox{Tree}{\tiny\strut \textcolor{white}{Tree}}%
\colorbox{Vegetation}{\tiny\strut Vegetation}%
\colorbox{Building}{\tiny\strut Building}%
\colorbox{Road}{\tiny\strut \textcolor{white}{Road}}%
\colorbox{GuardRail}{\tiny\strut GuardRail}%
\colorbox{TrafficSign}{\tiny\strut TrafficSign}%
\colorbox{TrafficLight}{\tiny\strut TrafficLight}%
\colorbox{Pole}{\tiny\strut \textcolor{white}{Pole}}%
\colorbox{Misc}{\tiny\strut Misc}%
\colorbox{Car}{\tiny\strut Car}%
\colorbox{Truck}{\tiny\strut \textcolor{white}{Truck}}%
\colorbox{Van}{\tiny\strut Van}
\ifx\showqualiresults\undefined
\else
\includegraphics[width=0.24\linewidth]{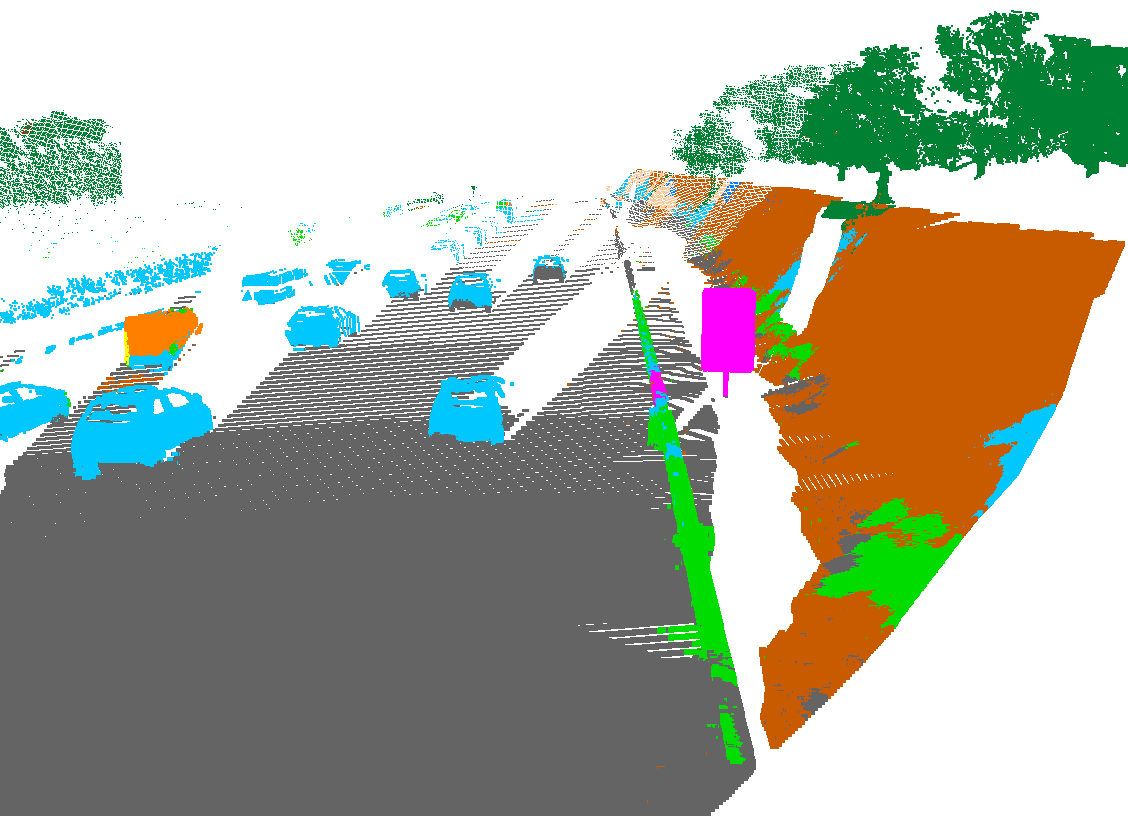} %
\includegraphics[width=0.24\linewidth]{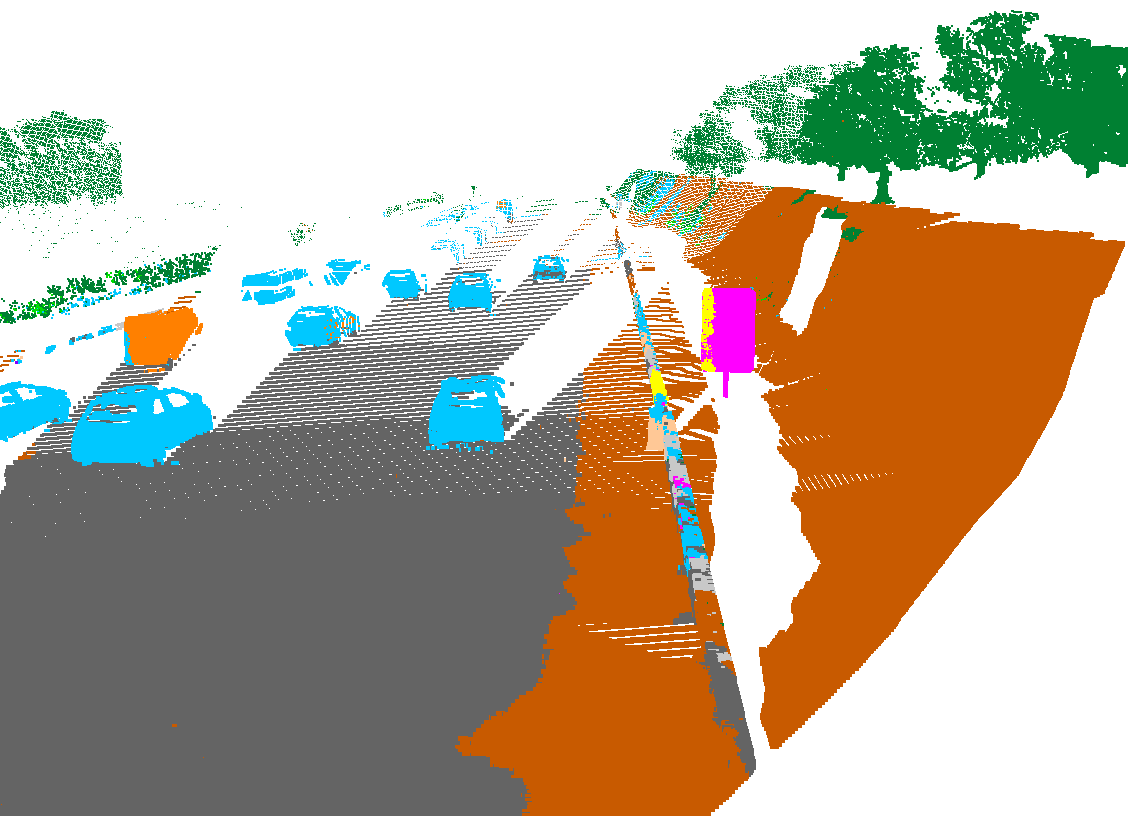} %
\includegraphics[width=0.24\linewidth]{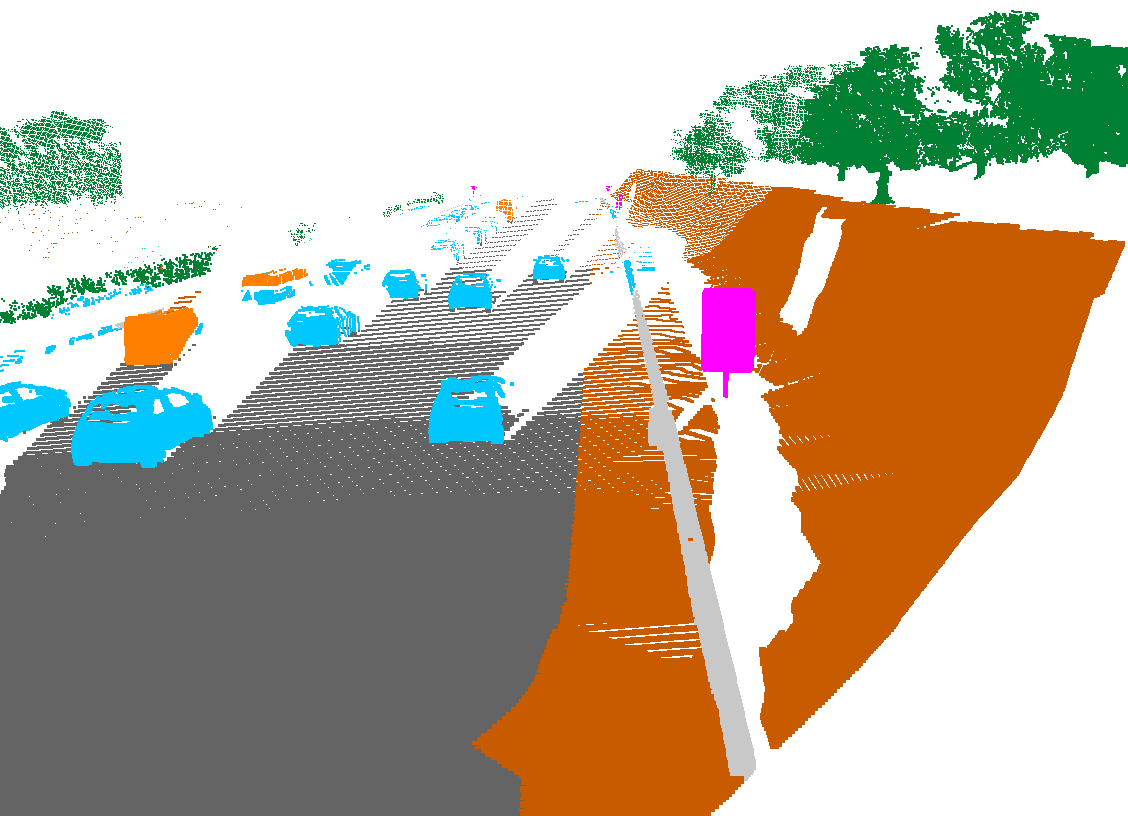} %
\includegraphics[width=0.24\linewidth]{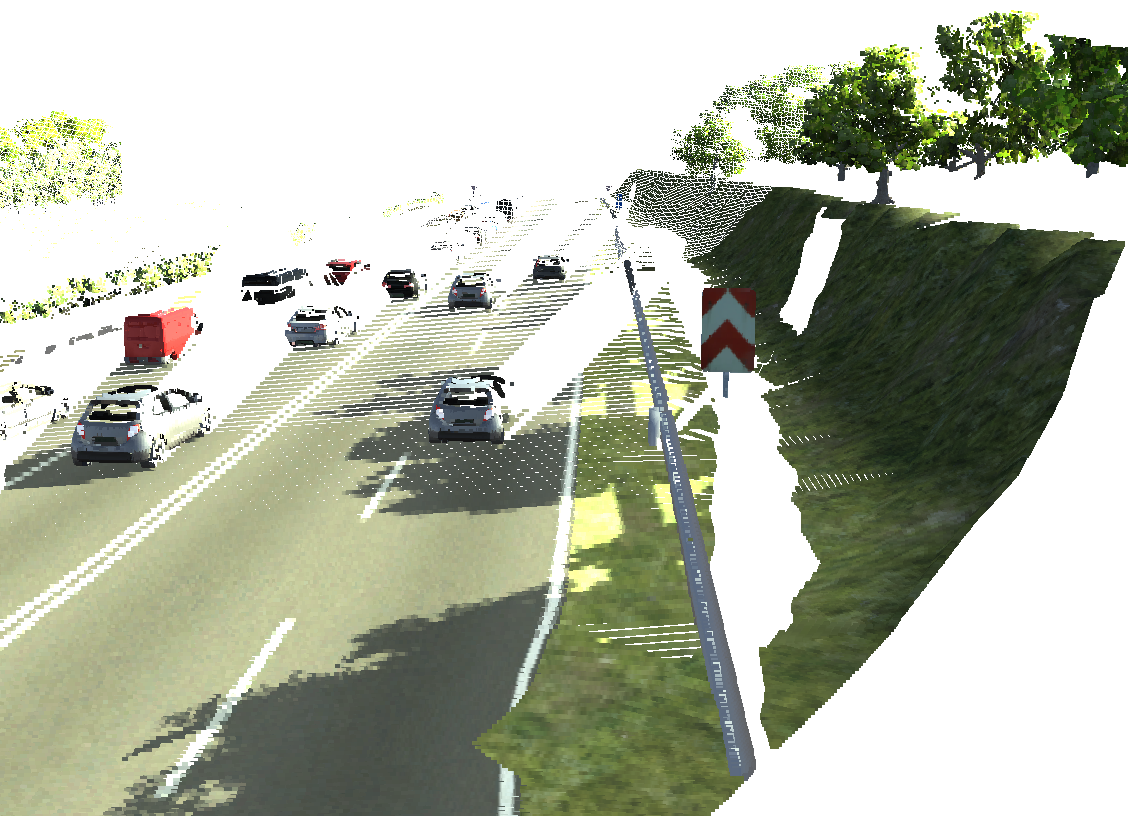} %

\includegraphics[width=0.24\linewidth]{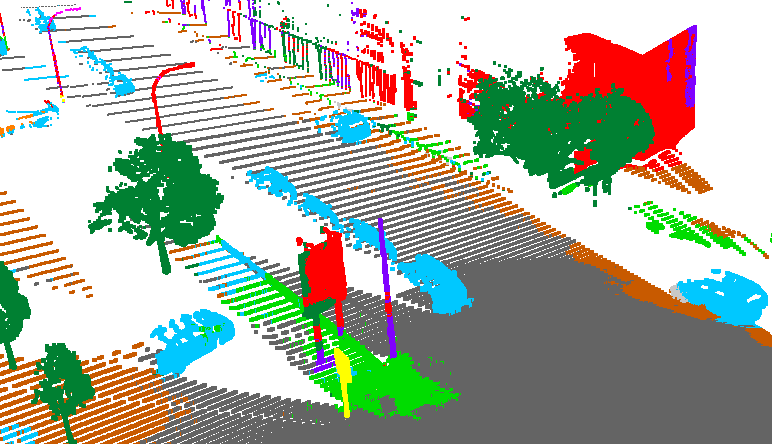} %
\includegraphics[width=0.24\linewidth]{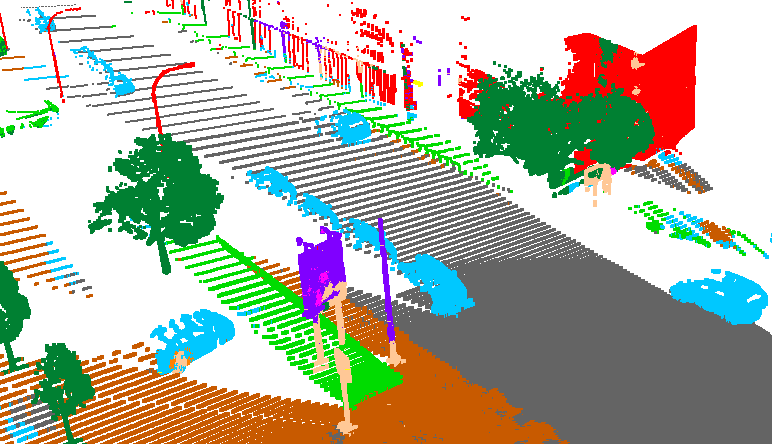} %
\includegraphics[width=0.24\linewidth]{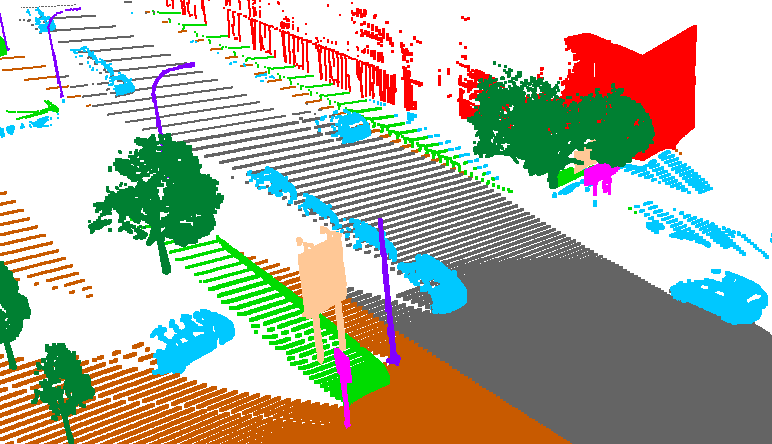} %
\includegraphics[width=0.24\linewidth]{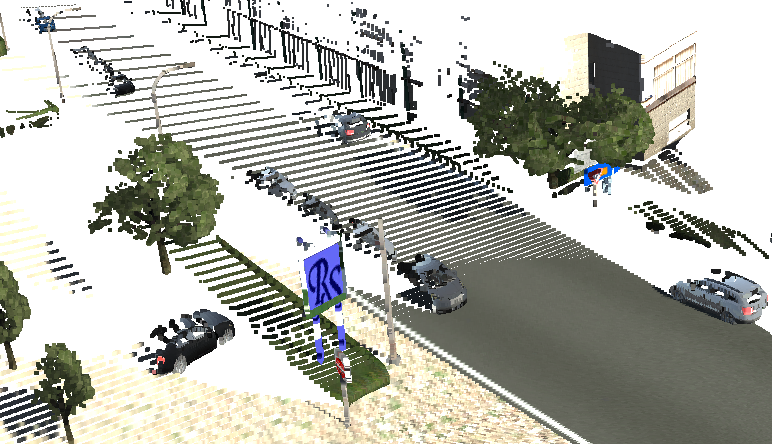} %

\includegraphics[width=0.24\linewidth]{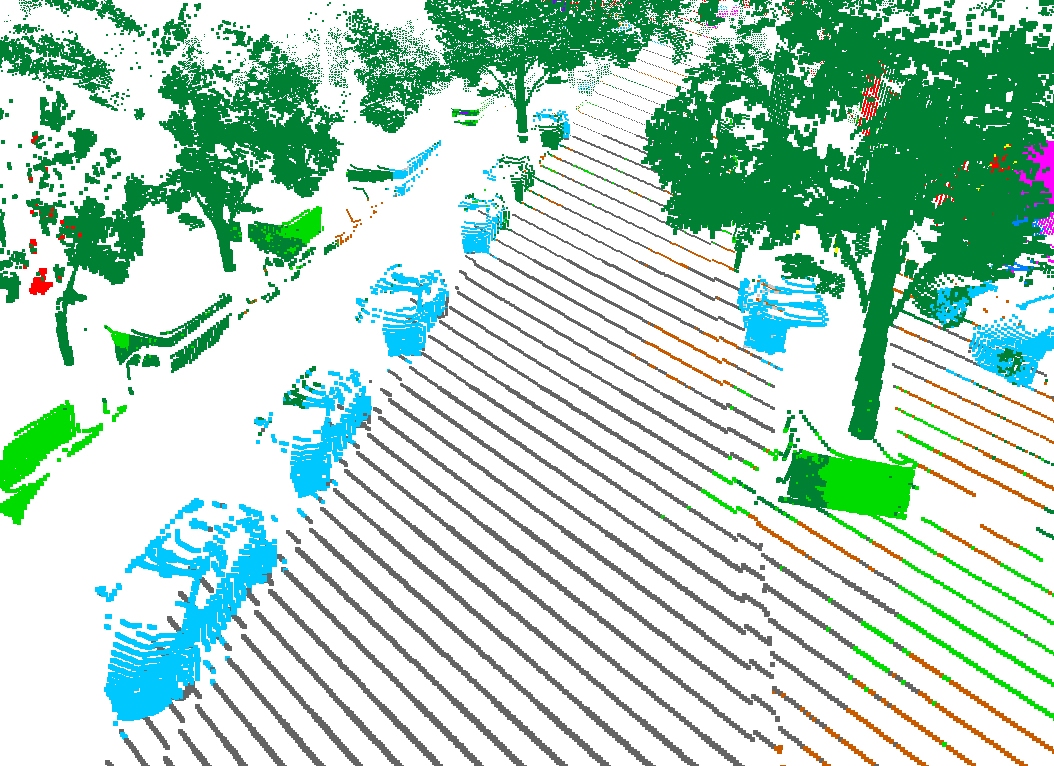} %
\includegraphics[width=0.24\linewidth]{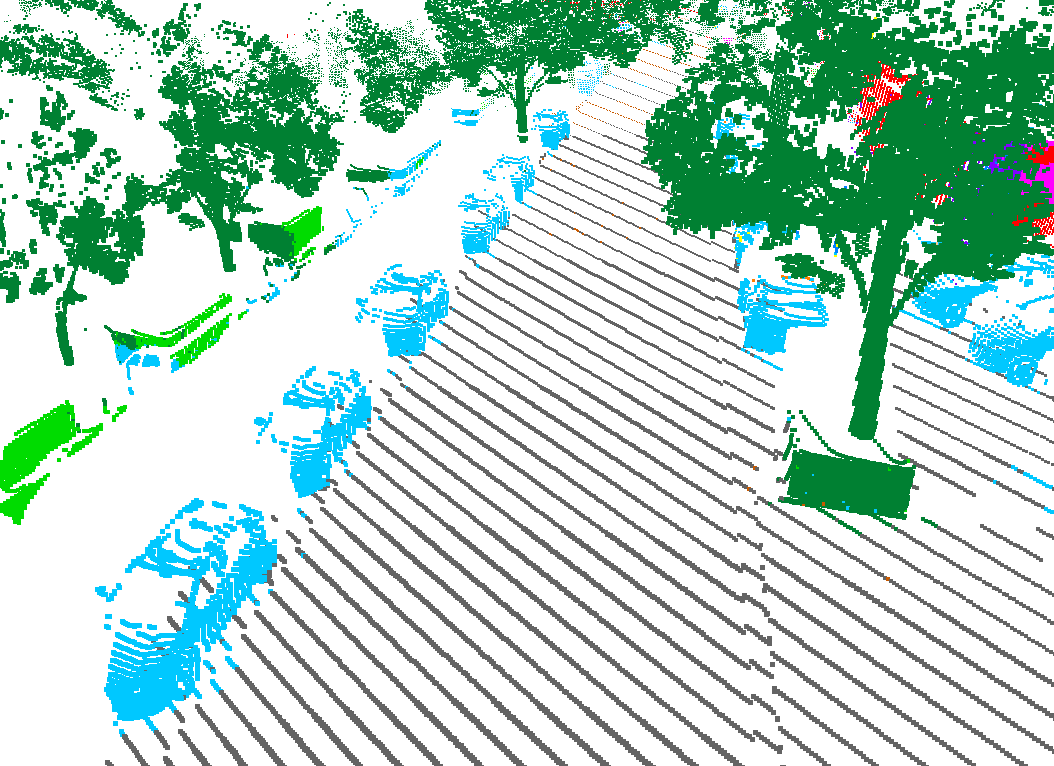} %
\includegraphics[width=0.24\linewidth]{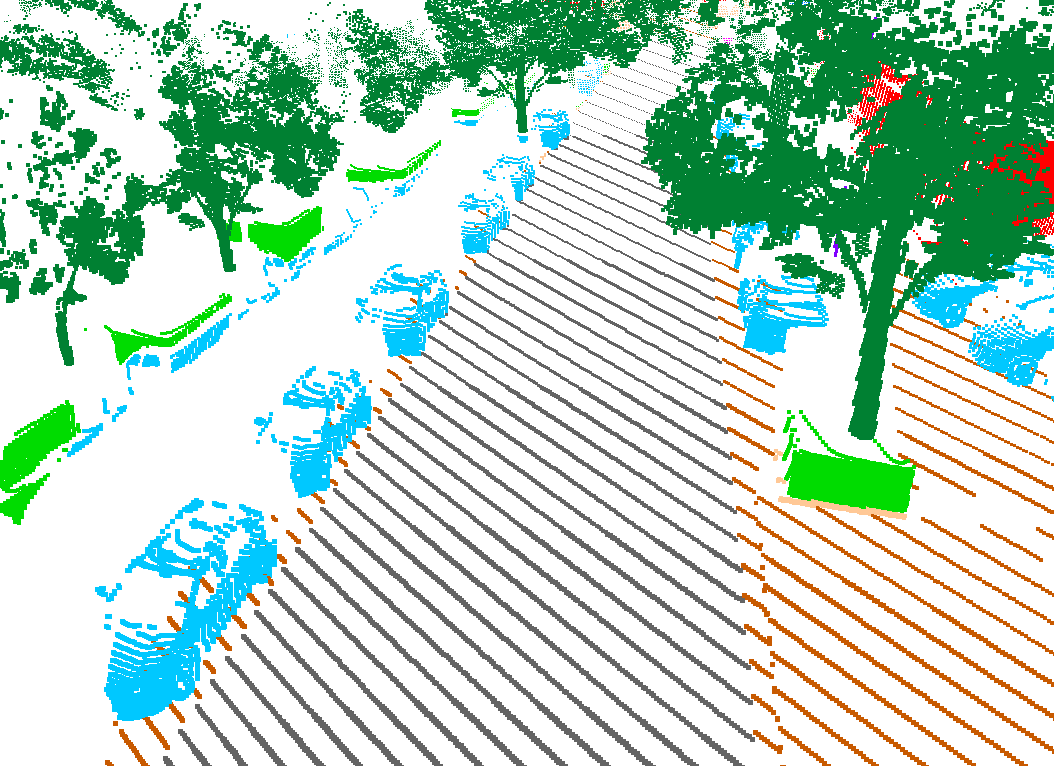} %
\includegraphics[width=0.24\linewidth]{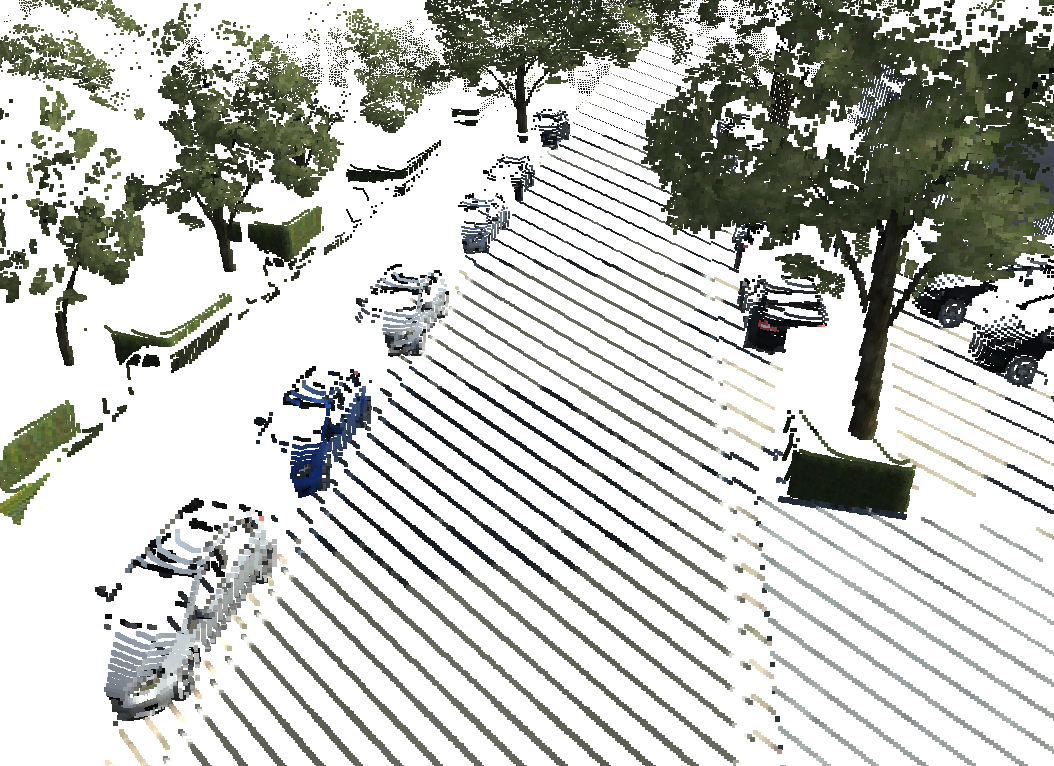} %

\includegraphics[width=0.24\linewidth]{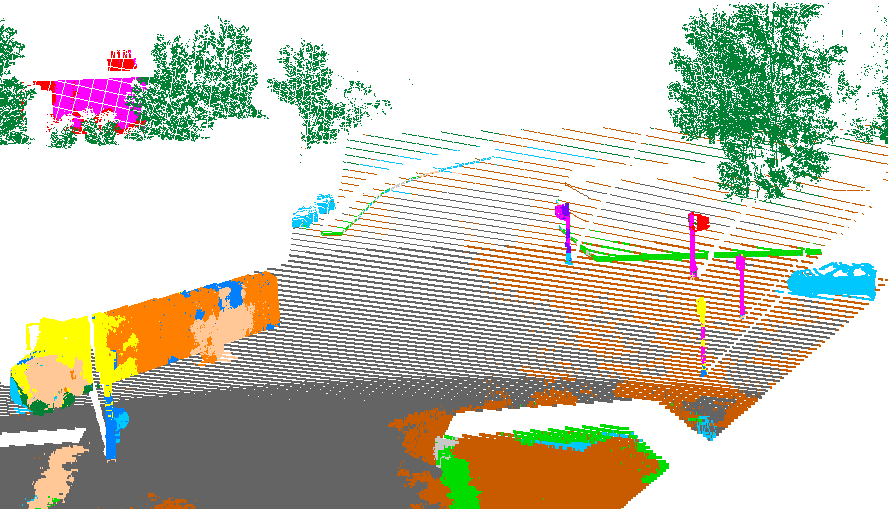} %
\includegraphics[width=0.24\linewidth]{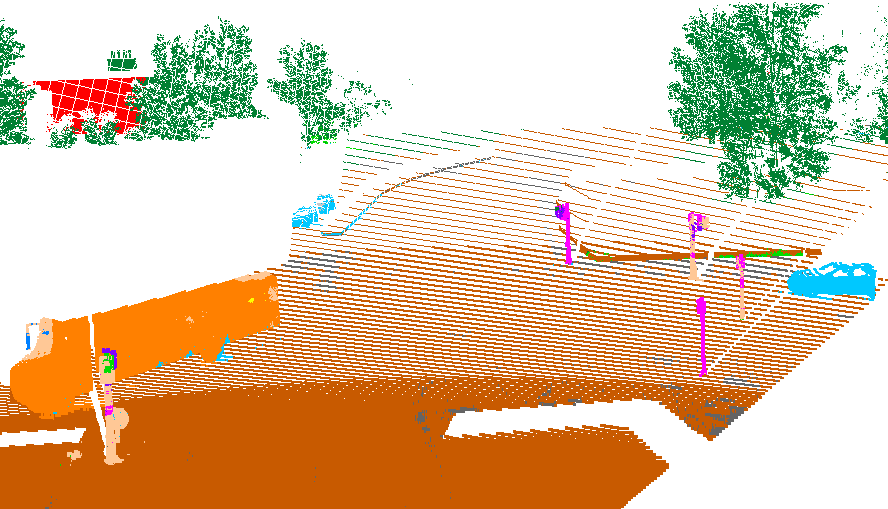} %
\includegraphics[width=0.24\linewidth]{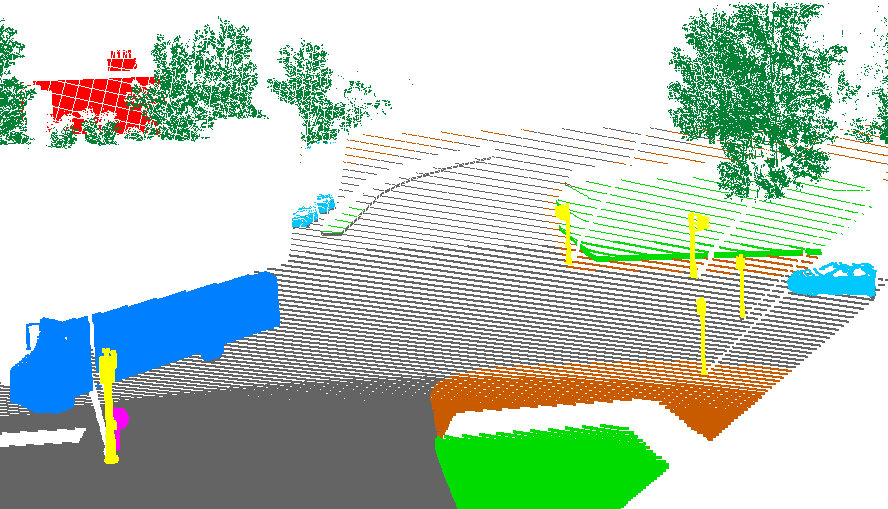} %
\includegraphics[width=0.24\linewidth]{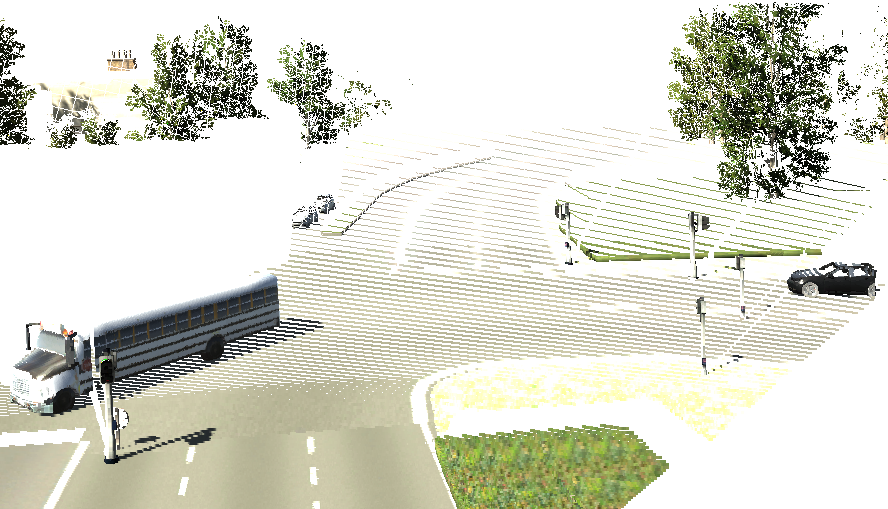} %
\fi

\begin{minipage}{0.24\textwidth}
\center
\textbf{\small Ours (No Color)}
\label{fig:no_color}
\end{minipage}
\begin{minipage}{0.24\textwidth}
\center
\textbf{\small Ours (Color)}
\label{fig:color}
\end{minipage}
\begin{minipage}{0.24\textwidth}
\center
\textbf{\small Ground Truth}
\label{fig:gt}
\end{minipage}
\begin{minipage}{0.24\textwidth}
\center
\textbf{\small RGB}\label{fig:rgb}
\end{minipage}
\caption{
	\label{fig:quali_results_vkitti3d}
	Qualitative results on VKITTI3D dataset. In general, color is an important attribute to distinguish between shapes that have similar structure e.g. 'terrain' and 'road'.
	The last row shows a failure case, during training our model was not able to differentiate between 'Van' and 'Truck', and between 'Terrain' and 'Road'.}
\end{figure*}

On the VKITTI3D dataset, we follow again the same training procedure as on the S3DIS dataset. The scenes are split into blocks of 9 m$^2$ on the ground plane.
Since the dataset is much sparser, we set the number of points sampled per block to N = 256.
Training on a 6-fold cross validation is performed as in \cite{us}.
We use the same input features as in the indoor dataset and additionally, we analyze how well our method performs if we take into consideration only geometric features (xyz-position) while leaving out color information.
This is an interesting experiment, as color is not always imminently available e.g. point clouds from laser scanners. 
We show quantitative results in \reftab{vkitti3d_quanti_short}.
Qualitative results are shown in \reffig{quali_results_vkitti3d}.

\begin{table}[t]
\caption{
\textbf{ScanNet.}
Overall point accuracy (oAcc), mean semantic class accuracy (mAcc), mean Intersection-over-Union (mIoU).
ScanNet dataset using the official training and test split from \cite{scannet}, scores are shown on a per-point basis as computed by the PN++ reference implementation.
To train on our hardware, we set the batch size to 32 and number of points to 1024}
\center
\begin{tabular}[t]{r ccc}
\toprule
\textbf{ScanNet} \cite{scannet} & \, oAcc \, & \, mAcc \, \\
\midrule
PointNet++\,\cite{pointnetpp}  &  71.40  & 24.51 \\
Our method & \textbf{75.53} & \textbf{25.39}  \\
\bottomrule
\end{tabular}
\label{tab:quanti_scannet}
\end{table}

\begin{table}[t]
\caption{Study on the feature network.
We evaluate the number of layers and compare feature fusion using concatenation or addition.
Deeper networks perform better, in general feature addition is slightly stronger while being more memory efficient than concatenation}
\center
\begin{tabular}[b]{ccc}
\toprule
\# Layers & \, Fusion \, & \, mIoU \, \\
\midrule
3 & additive & 42.15 \\
12 & additive & 44.46 \\
17 & additive & \textbf{45.15} \\ 
17 & concat. & 44.23 \\ 
12 & concat. & 43.35 \\
3  & concat. & 41.73 \\
\bottomrule
\end{tabular}
\label{tab:feature_network_eval}
\end{table}
\section{Conclusion}
We have presented a deep learning framework for 3D semantic segmentation of point clouds.
Its main components are $\mathcal{N}_F$- and $\mathcal{N}_W$-modules. 
They allow to incorporate neighborhood information from the feature space and from the world space.
We have also introduced the pairwise distance loss $\mathcal{L}_{pair}$ and the centroid loss $\mathcal{L}_{cent}$ in the context of 3D semantic segmentation.
The presented method produces state-of-the-art results on current indoor and outdoor datasets.
\subsection*{Acknowledgement}
This project was funded by the ERC Consolidator Grant DeeViSe (ERC-2017-CoG-773161).

%

%

\clearpage
\vspace{-10cm}
\vspace{-1cm}
\begin{table}[h!]
\caption{\textbf{Virtual KITTI 3D.}
The upper part of the tables shows results trained on position only.
In the lower part, we additionally trained with color.
Geometric features alone are quite powerful.
Adding color helps to differ between geometric similar classes}
\center
\begin{tabular}{rccc}
\toprule
\textbf{VKITTI3D \cite{us}: 6-fold CV} & \, oAcc \, & \, mAcc \, & \, mIoU \, \\
\midrule
PointNet \cite{pointnet} from \cite{us} & 63.3 & 29.9 & 17.9\\
MS+CU(2) \cite{us} 				 & 73.2 & 40.9 & 26.4\\
Ours                              			 & \textbf{78.19} & \textbf{56.43} & \textbf{33.36}\\
\midrule
Ours ( + color)					 & \textbf{79.69} & \textbf{57.59} & \textbf{35.59} \\
\bottomrule
\end{tabular}
\label{tab:vkitti3d_quanti_short}
\end{table}

\clearpage
%
%
%
\bibliographystyle{splncs04}
\bibliography{egbib}

\end{document}